%% file: main.tex
\documentclass[10pt,twocolumn,letterpaper]{article}

\usepackage[pagenumbers]{cvpr} 

\input{preamble}

\definecolor{cvprblue}{rgb}{0.21,0.49,0.74}
\usepackage[pagebackref,breaklinks,colorlinks,allcolors=cvprblue]{hyperref}

\def\paperID{4542} 
\def\confName{CVPR}
\def\confYear{2025}

\title{Nearly Zero-Cost Protection Against Mimicry by Personalized Diffusion Models}

\author{
    Namhyuk Ahn $^{1,2}$ $\quad$
    KiYoon Yoo $^3$ $\quad$
    Wonhyuk Ahn $^2$ $\quad$
    Daesik Kim $^2$ $\quad$
    Seung-Hun Nam $^2$  \\
    $^1$ Inha University \quad $^2$ NAVER WEBTOON AI \quad $^3$ KRAFTON \\
}

\begin{document}

\twocolumn[{
\renewcommand\twocolumn[1][]{#1}
\maketitle
\begin{center}
\centering
\captionsetup{type=figure}
\vspace{-0.5em}
\centering
\begin{subfigure}[b]{0.48\textwidth}
    \centering
    \includegraphics[width=\linewidth]{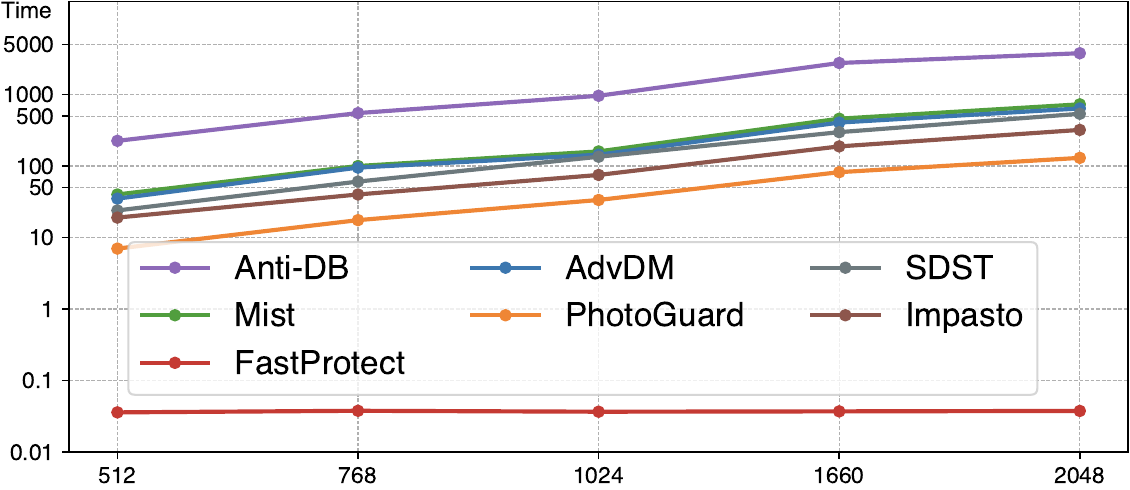}
    \caption{Inference latency (log-scaled) vs. image size}
    \label{fig:teaser_latency}
\end{subfigure}
\hfill
\begin{subfigure}[b]{0.495\textwidth}
    \centering
    \includegraphics[width=\linewidth]{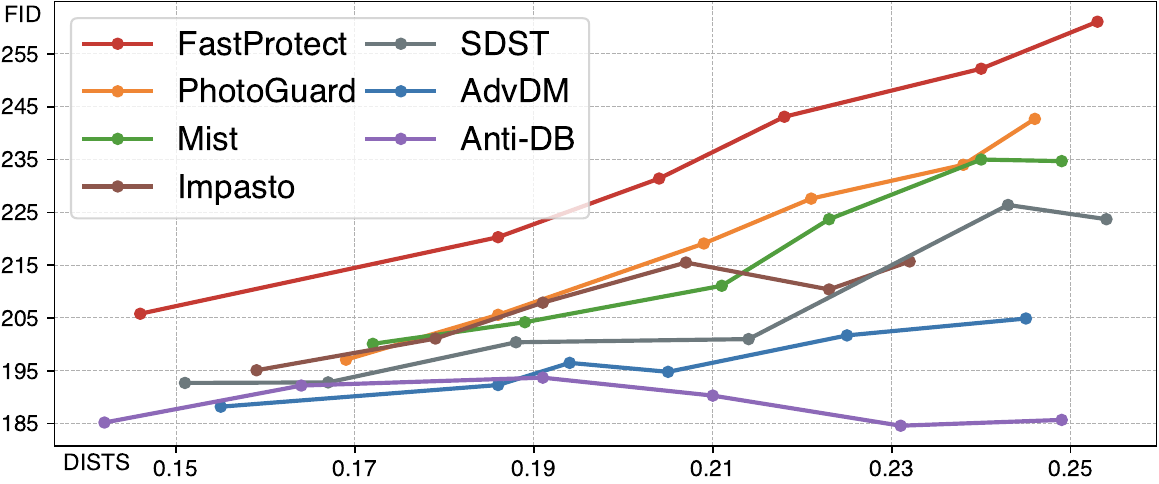}
    \caption{Protection efficacy vs. invisibility}
    \label{fig:teaser_tradeoff}
\end{subfigure}
\hfill
\vspace{-0.5em}
\caption{(a) \ours\ shows unprecedented speed in protection against diffusion models. On an A100 GPU, \ours\ achieves real-time latency even for processing $2048^2$-px image, while others require substantially longer time.
(b) In terms of the trade-off between protection efficacy (FID, $\uparrow$ is better) and invisibility (DISTS, $\downarrow$ is better), \ours\ exhibits improvement over other protection methods.}
\label{fig:teaser}
\end{center}
}]

\input{sections/0_abstract}

\input{sections/1_introduction}
\input{sections/2_background}

\input{sections/3_method}
\input{sections/4_experiments}
\input{sections/5_conclusion}

\clearpage

\noindent\textbf{Acknowledgment.} 
This work was partly supported by NAVER WEBTOON and by Institute of Information \& communications Technology Planning \& Evaluation (IITP) grant funded by the Korea government(MSIT) under the Artificial Intelligence Convergence Innovation Human Resources Development, Inha University (No.RS-2022-00155915) and under the Leading Generative AI Human Resources Development
(IITP-2025-RS-2024-00360227).

{
    \small
    \bibliographystyle{ieeenat_fullname}
    \bibliography{main}
}

\clearpage
\appendix
\input{sections/6_appendix}

\end{document}

%% file: preamble.tex
\usepackage[dvipsnames]{xcolor}

\usepackage{multirow}
\usepackage[ruled]{algorithm2e}

\input{sections/figures}
\input{sections/tables}
\input{sections/algorithms}
\newcommand{\ours}{FastProtect}
\newcommand{\suppl}{Appendix}

\newcommand{\sref}[1]{Section~\ref{#1}}
\newcommand{\eref}[1]{Eq.~\ref{#1}}
\newcommand{\tref}[1]{Table~\ref{#1}}
\newcommand{\fref}[1]{Figure~\ref{#1}}
\newcommand{\aref}[1]{Algorithm~\ref{#1}}

\DeclareMathOperator*{\argmin}{arg\,min}
\DeclareMathOperator*{\argmax}{arg\,max}

\definecolor{cadmiumgreen}{rgb}{0.0, 0.42, 0.24}

\SetCommentSty{mycommfont}

%% file: sections/figures.tex
\newcommand{\figModelOverview}{
\begin{figure*}[t]
\centering
\includegraphics[width=\linewidth]{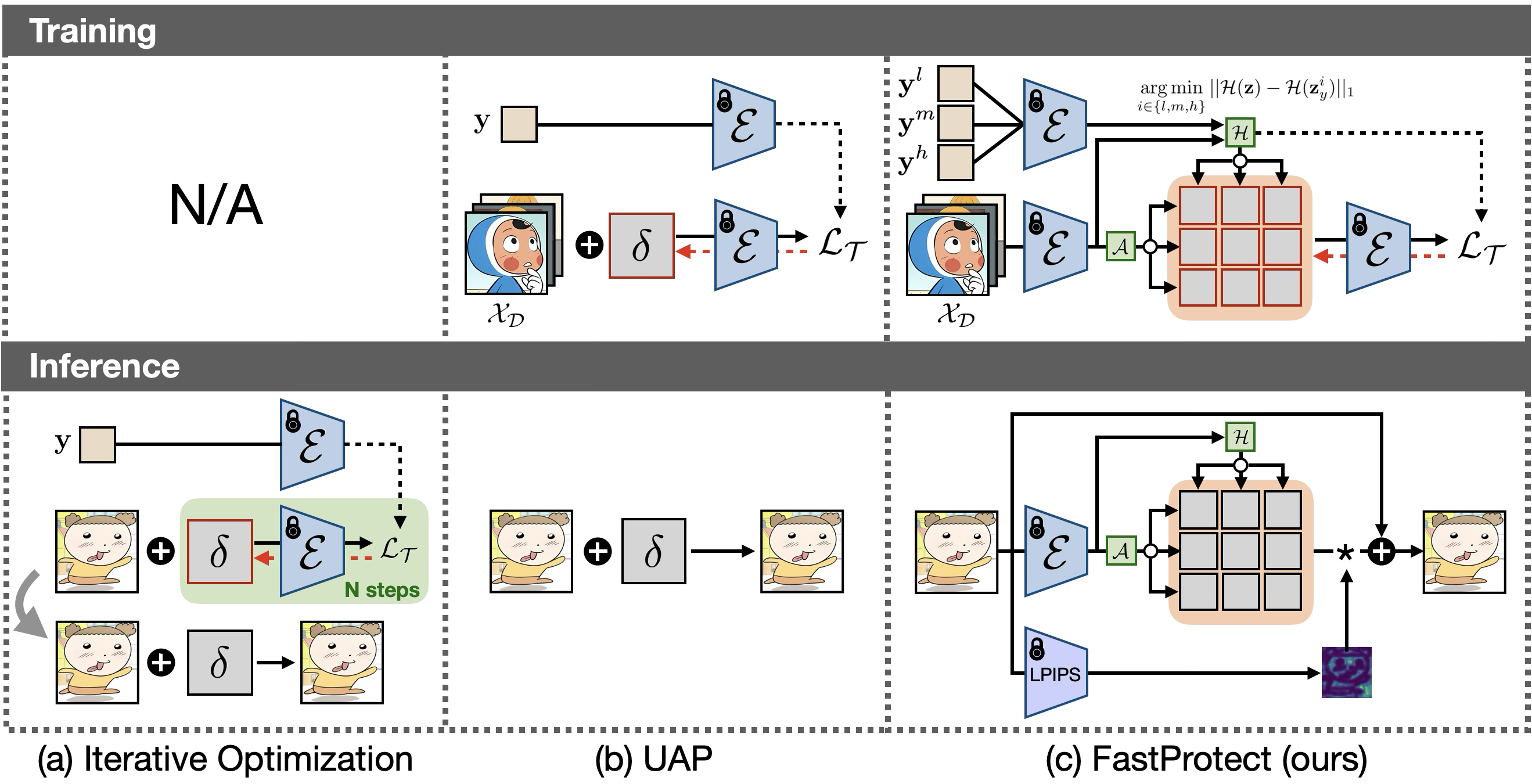}
\vspace{-1.5em}
\caption{
Model overview. (a) Current iterative optimization approaches lack a training phase and perform optimization during inference, resulting in extremely slow protection. (b) UAP~\cite{moosavi2017universal} introduces pre-training of perturbations, but their image-agnostic nature leads to degraded protection efficacy. (c) Combining the advantages of both paradigms, \ours\ adopts a pre-training approach similar to UAP but with a novel mixture-of-perturbation scheme and multi-layer protection loss to enhance protection efficacy. At inference, adaptive targeted protection further boosts protection efficacy with minimal additional cost, and adaptive protection strength improves invisibility.
}
\vspace{-1em}
\label{fig:model_overview}
\end{figure*}
}

\newcommand{\figTarget}{
\begin{figure}[t]
\centering
\includegraphics[width=\linewidth]{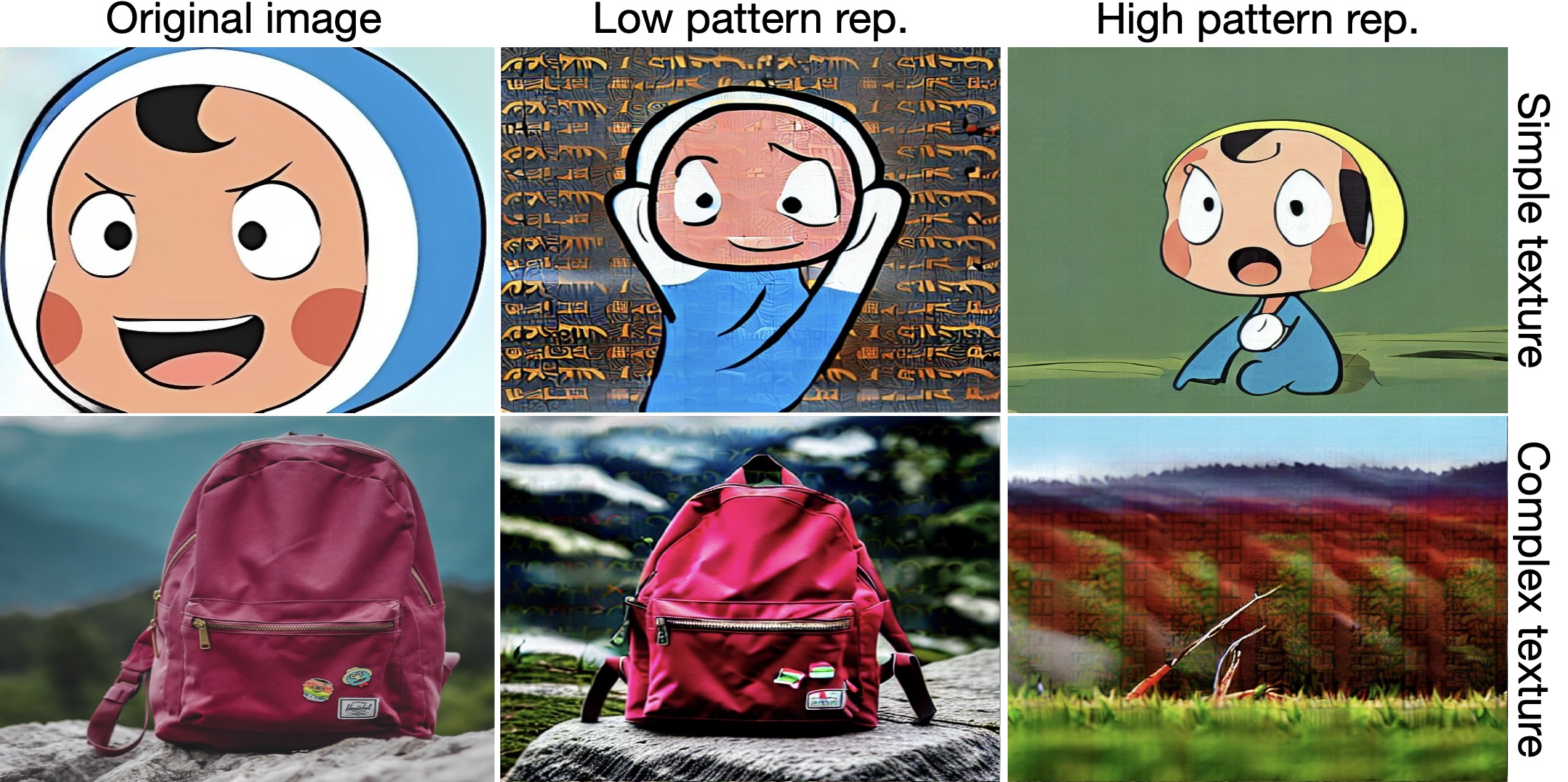}
\vspace{-1.5em}
\caption{Relationship between target image's pattern repetition and input image’s texture. Simple textured image is successfully protected by a low repetition target, but fails when using a high repetition target; vice versa for complex texture cases.}
\label{fig:target_observation}
\vspace{-1em}
\end{figure}
}

\newcommand{\figLPIPS}{
\begin{figure}[t]
\centering
\includegraphics[width=\linewidth]{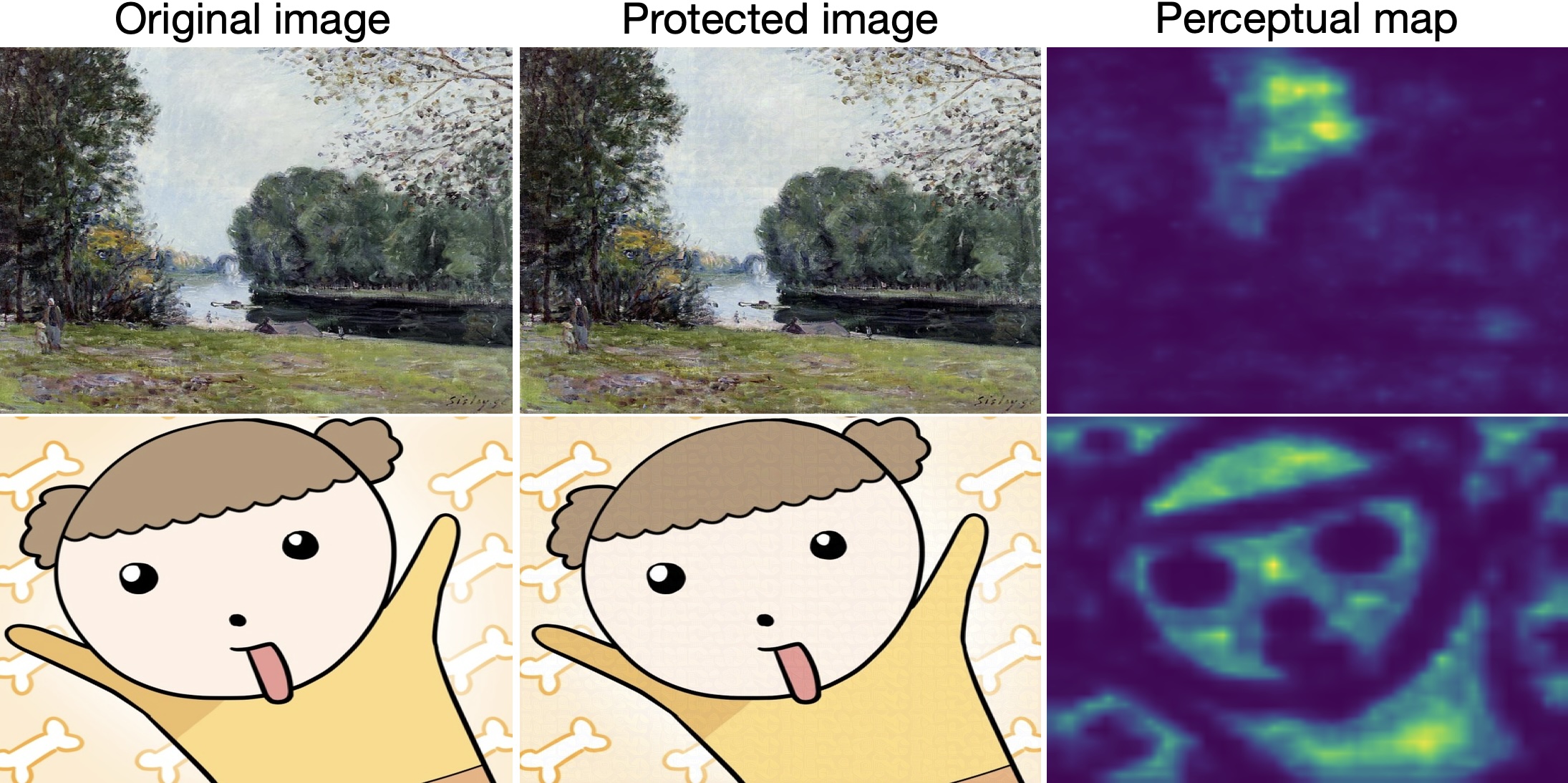}
\vspace{-1.5em}
\caption{Given the original and protected images, we obtain the LPIPS distance map, which remarkably aligns with human perception. The brighter regions on the perceptual map indicate areas where subtle distortions are more noticeable.}
\label{fig:lpips_map}
\vspace{-1em}
\end{figure}
}

\newcommand{\figQual}{
\begin{figure*}[t]
\centering
\includegraphics[width=0.99\linewidth]{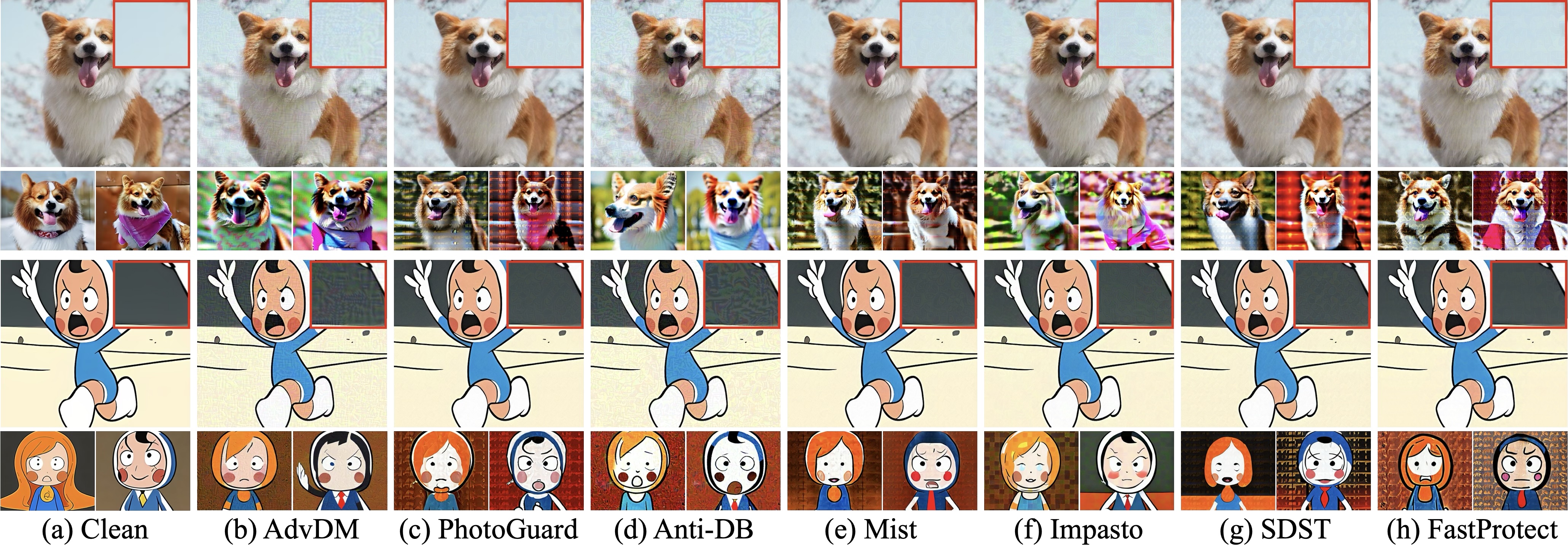}
\vspace{-0.5em}
\caption{Qualitative comparison of different protection frameworks. (Top) Protected image with a zoomed-in patch in the inset. (Bottom) Two output images from the personalized LoRA.}
\label{fig:qual}
\vspace{-1.em}
\end{figure*}
}

\newcommand{\figInferenceAnalysis}{
\begin{figure*}[t]
\centering
\begin{subfigure}[b]{0.32\textwidth}
    \centering
    \includegraphics[width=\linewidth]{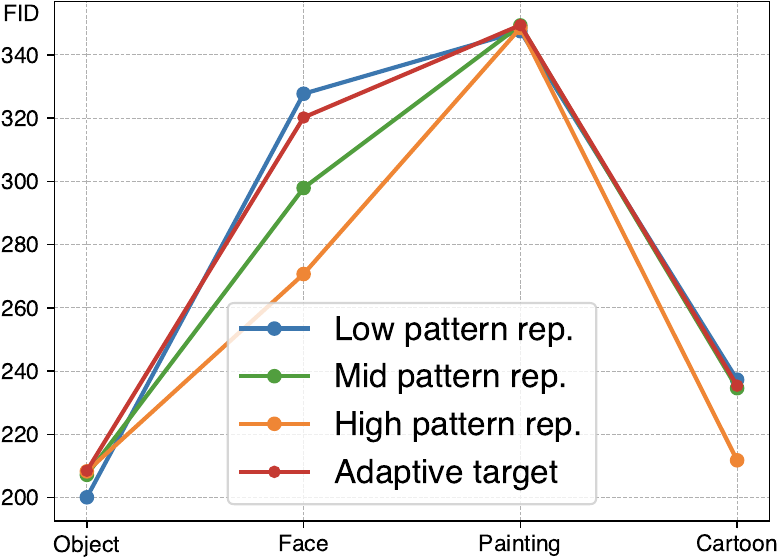}
    \caption{Adapt. Targeted Protection}
    \label{fig:analysis_target}
\end{subfigure}
\hfill
\begin{subfigure}[b]{0.32\textwidth}
    \centering
    \includegraphics[width=\linewidth]{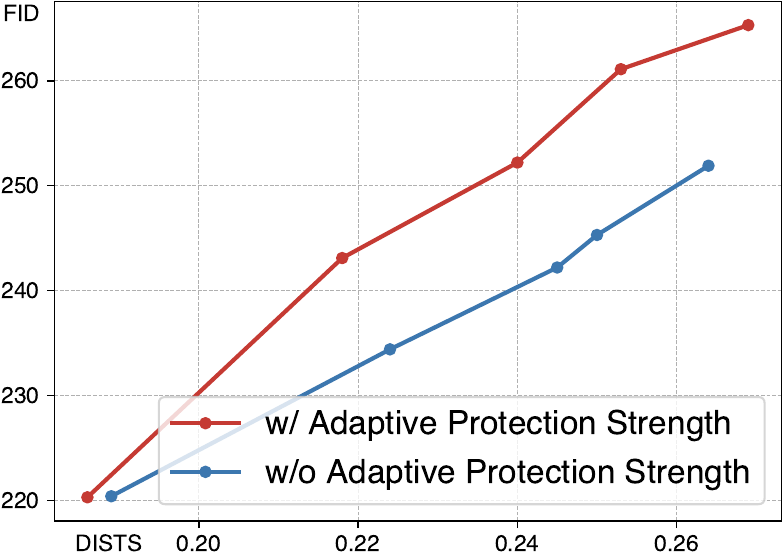}
    \caption{Adapt. Protection Strength}
    \label{fig:analysis_lpips}
\end{subfigure}
\hfill
\begin{subfigure}[b]{0.32\textwidth}
    \centering
    \includegraphics[width=\linewidth]{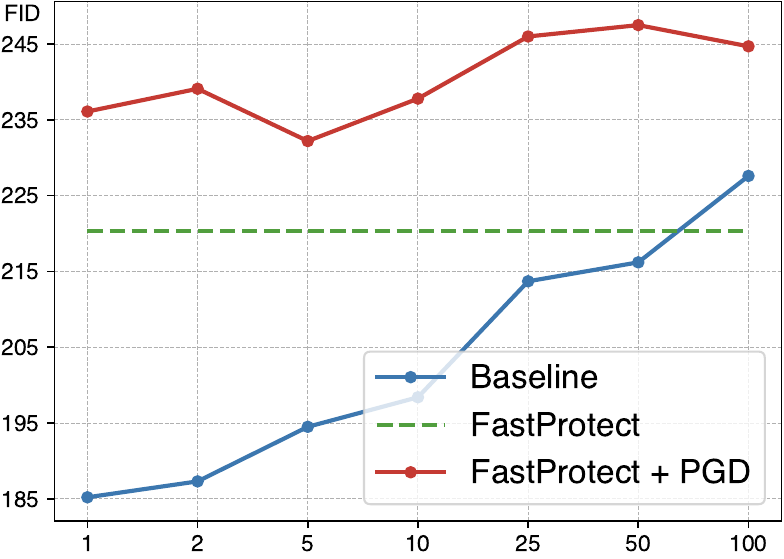}
    \caption{\ours\ + PGD Refine}
    \label{fig:analysis_add_pgd}
\end{subfigure}
\hfill
\vspace{-0.4em}
\caption{Analysis of the proposed modules in the inference phase.}
\label{fig:inference_analysis}
\vspace{-1em}
\end{figure*}
}

\newcommand{\figTrainAnalysis}{
\begin{figure*}[t]
\centering
\begin{subfigure}[b]{0.32\textwidth}
    \centering
    \includegraphics[width=\linewidth]{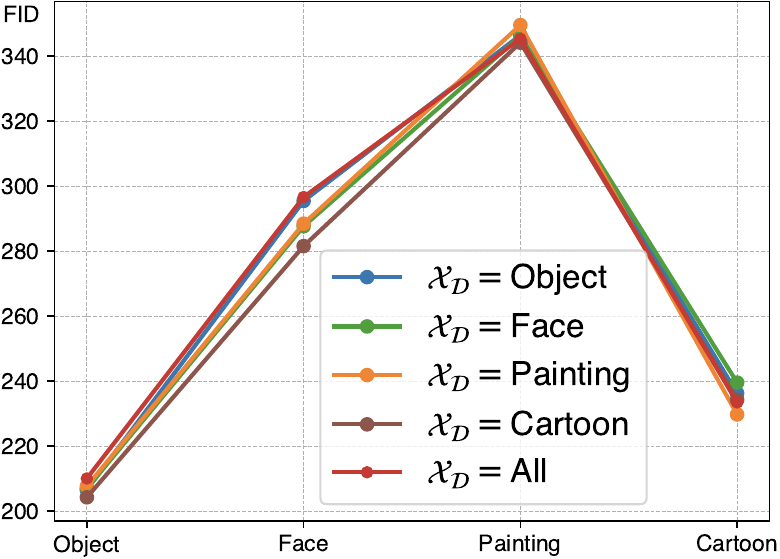}
    \caption{Domain Generalization of MoP}
    \label{fig:analysis_domain}
\end{subfigure}
\hfill
\begin{subfigure}[b]{0.32\textwidth}
    \centering
    \includegraphics[width=\linewidth]{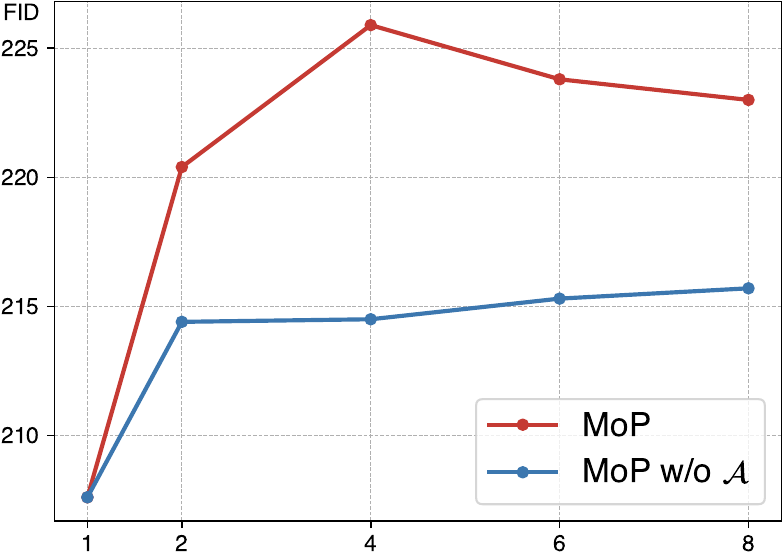}
    \caption{Effect of Assignment Function}
    \label{fig:analysis_k}
\end{subfigure}
\hfill
\begin{subfigure}[b]{0.32\textwidth}
    \centering
    \includegraphics[width=\linewidth, height=3.35cm]{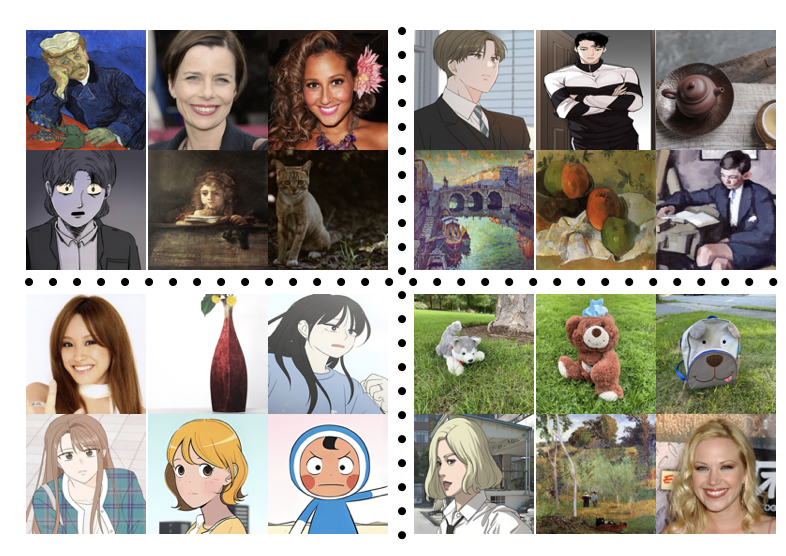}
    \caption{Examples of MoP Group}
    \label{fig:analysis_group}
\end{subfigure}
\hfill
\vspace{-0.4em}
\caption{Analysis of the proposed modules in the pre-training phase.}
\label{fig:train_analysis}
\end{figure*}
}

\newcommand{\figAppendixLatency}{
\begin{figure*}[t]
\centering
\begin{subfigure}[b]{0.48\textwidth}
    \centering
    \includegraphics[width=\linewidth]{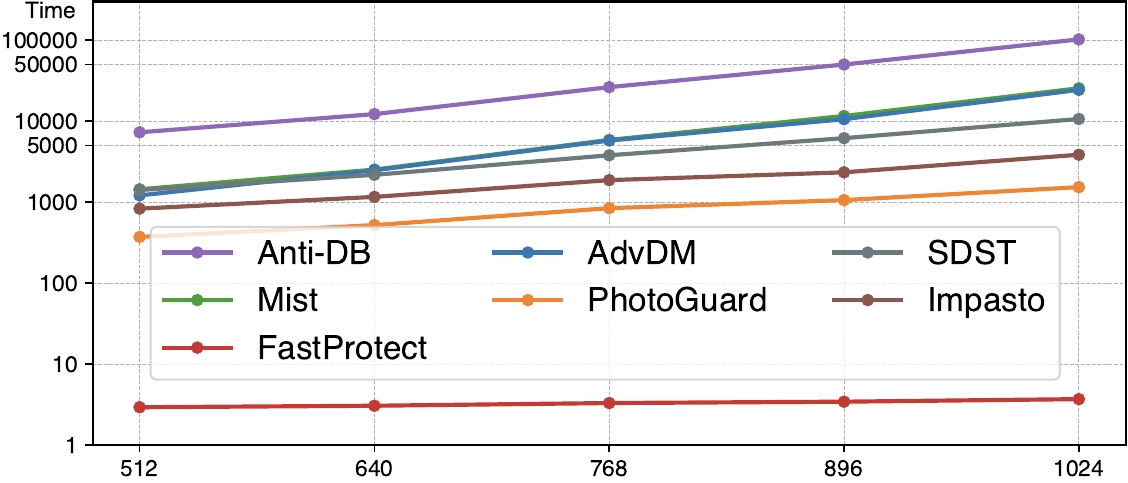}
    \caption{On CPU (Apple M1 Max)}
\end{subfigure}
\hfill
\begin{subfigure}[b]{0.48\textwidth}
    \centering
    \includegraphics[width=\linewidth]{figures/teaser/fig_latency_gpu.pdf}
    \caption{On GPU (NVIDIA A100)}
\end{subfigure}
\caption{Inference latency (log-scaled) vs. image size on both CPU and GPU environments.}
\label{fig:appendix_latency}
\vspace{-0.5em}
\end{figure*}
}

\newcommand{\figAppendixTradeoff}{
\begin{figure*}[t]
\centering
\begin{subfigure}[b]{0.48\textwidth}
    \centering
    \includegraphics[width=\linewidth]{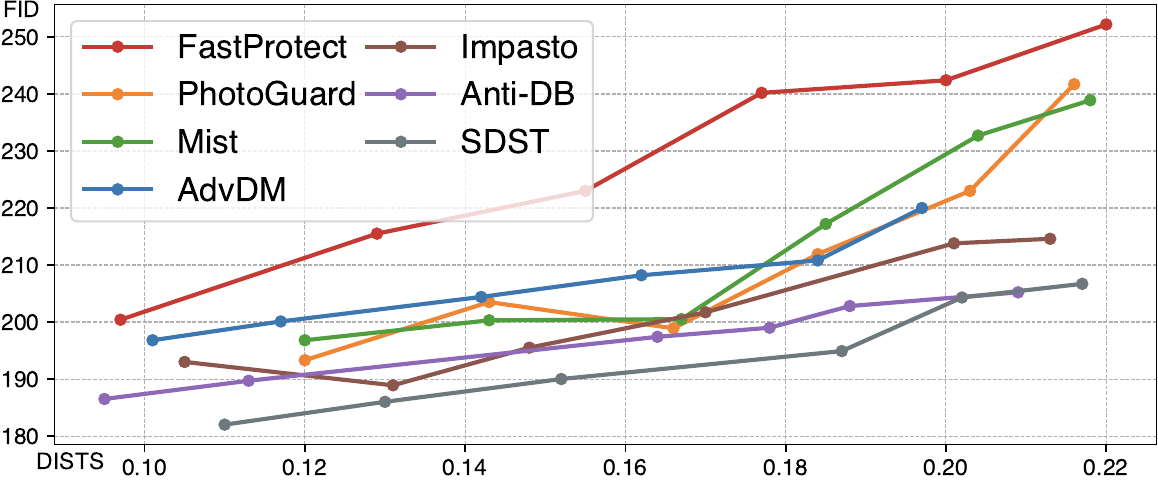}
    \caption{Object domain}
\end{subfigure}
\hfill
\begin{subfigure}[b]{0.48\textwidth}
    \centering
    \includegraphics[width=\linewidth]{figures/teaser/fig_tradeoff_webtoon.pdf}
    \caption{Cartoon domain}
\end{subfigure}
\caption{Protection efficacy vs. invisibility comparison on the Object and Cartoon domains.}
\label{fig:appendix_tradeoff}
\vspace{-1em}
\end{figure*}
}

\newcommand{\figTargetExample}{
\begin{figure}[t]
\centering
\newcommand{\w}{26.6mm}
\newcommand{\wpad}{0mm}
\includegraphics[width=\w]{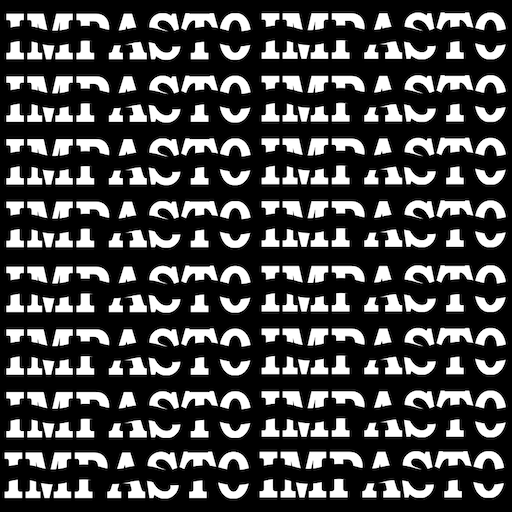}\hspace{\wpad}
\includegraphics[width=\w]{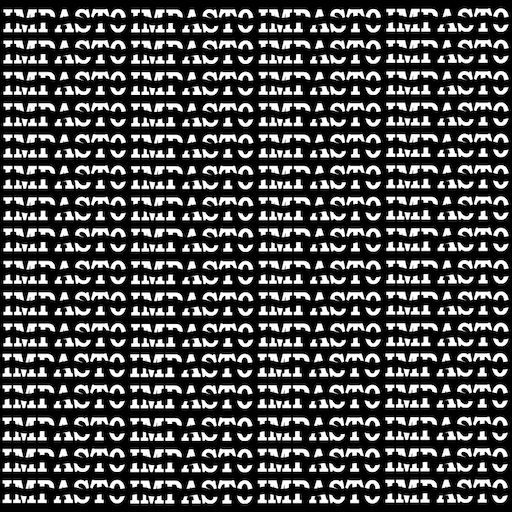}\hspace{\wpad}
\includegraphics[width=\w]{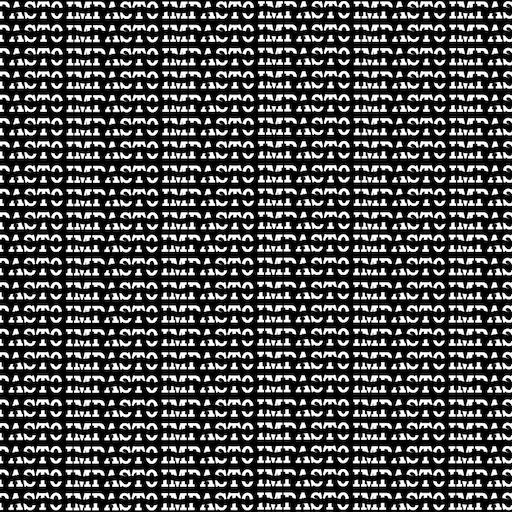 }\hfill
\\
\makebox[\w][c]{Low repetition}\hspace{\wpad}
\makebox[\w][c]{Mid repetition}\hspace{\wpad}
\makebox[\w][c]{High repetition}\hfill
\caption{Examples of target images used by \ours. In our target image analysis (\fref{fig:target_observation}), we used both low and high repetition target images.}
\label{fig:target_example}
\vspace{-0.5em}
\end{figure}
}

\newcommand{\figQualTwo}{
\begin{figure*}[t]
\centering
\includegraphics[width=\linewidth]{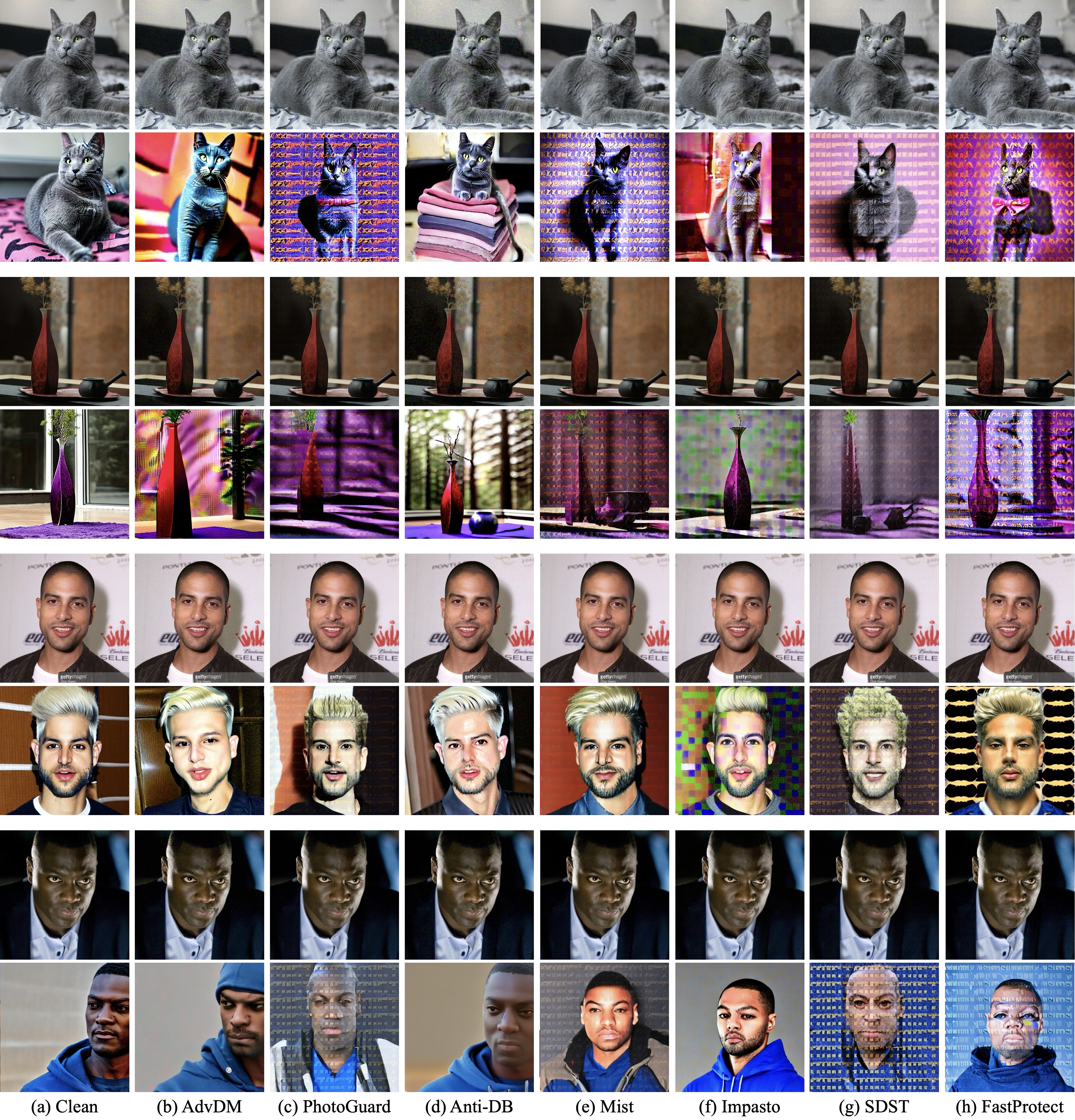}
\caption{Additional qualitative comparison. For each example, (top) protected image and (bottom) mimicry image generated by LoRA.}
\label{fig:qual2}
\vspace{-1em}
\end{figure*}
}

\newcommand{\figQualThree}{
\begin{figure*}[t]
\centering
\includegraphics[width=\linewidth]{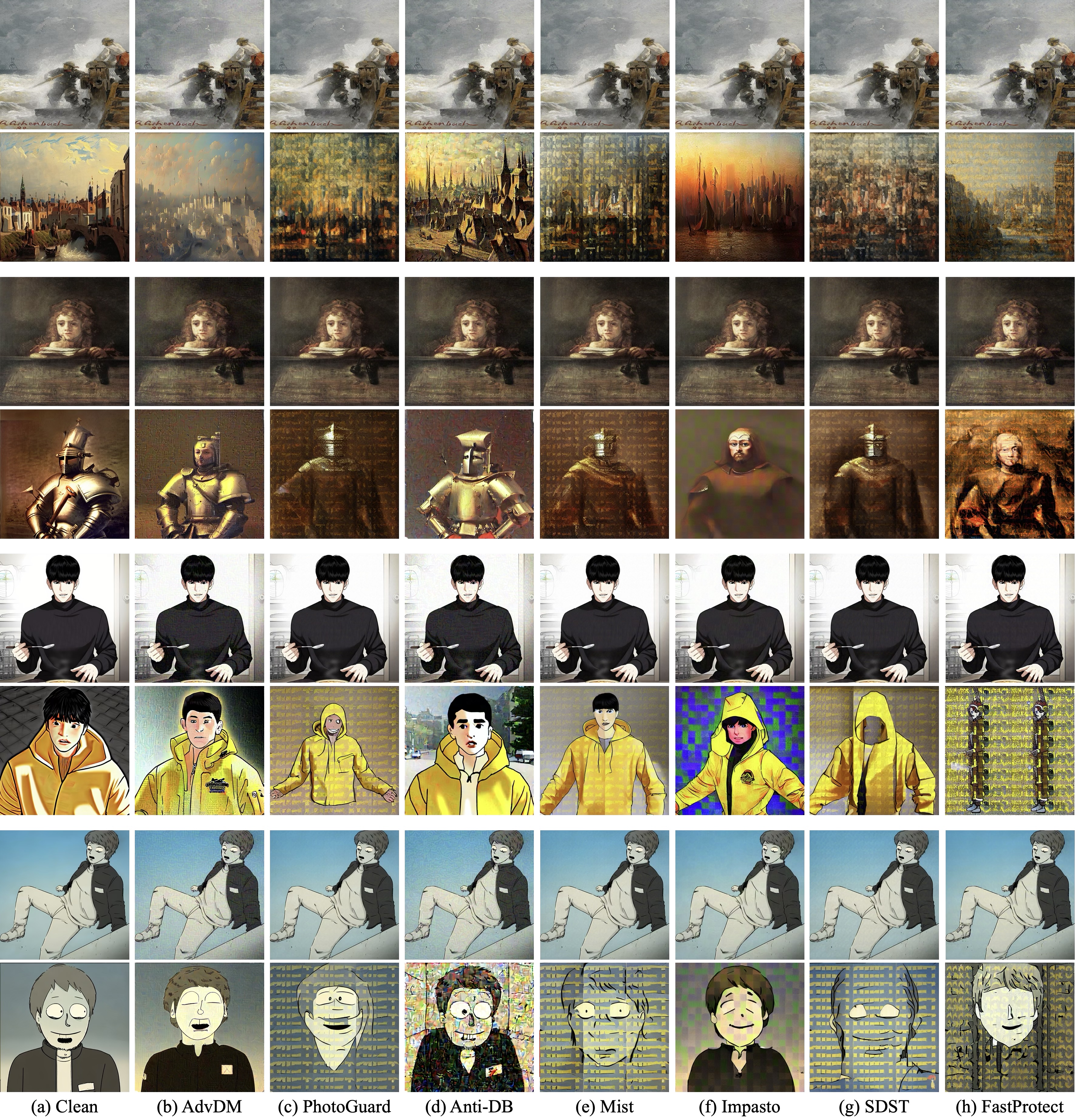}
\caption{Additional qualitative comparison. For each example, (top) protected image and (bottom) mimicry image generated by LoRA.}
\label{fig:qual3}
\vspace{-1em}
\end{figure*}
}

%% file: sections/tables.tex
\newcommand{\tablePGDUAP}{
\begin{table}[t]
\centering
\begin{minipage}[b]{\linewidth}
\centering
\caption{PGD vs. UAP (Top) Invisibility (DISTS) and protection efficacy (FID) comparison. (Bottom) Mimicry results via LoRA.}
\label{table:pgd_uap}
\vspace{-0.7em}
\setlength\tabcolsep{4pt}
\begin{tabular}{c|cc}
\hline
Method & Invisibility (DISTS; $\downarrow$) & Protection (FID; $\uparrow$) \\
\hline\hline
PGD & 0.221 & 227.6 \\
UAP & 0.222 & 207.6 \\
\hline
\end{tabular}
\end{minipage}%
\hfill
\vspace{0.1em}
\begin{minipage}[b]{0.45\textwidth}
\centering
\newcommand{\w}{25.6mm}
\newcommand{\h}{22.6mm}
\includegraphics[width=\w, height=\h]{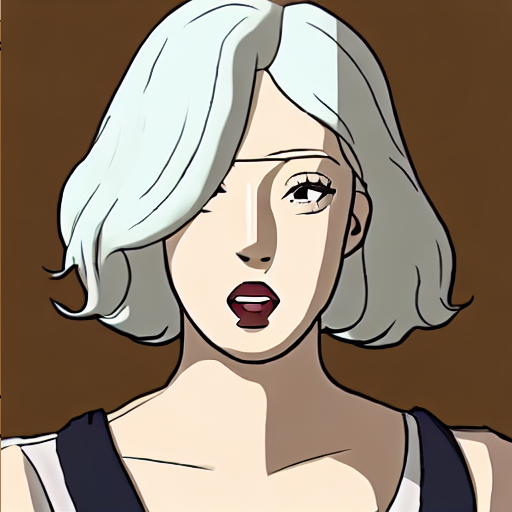}
\includegraphics[width=\w, height=\h]{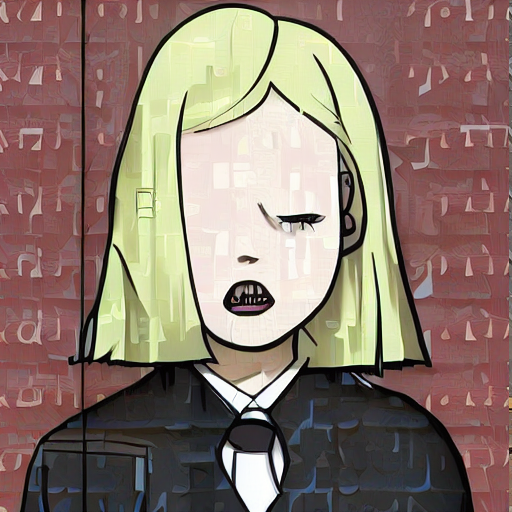}
\includegraphics[width=\w, height=\h]{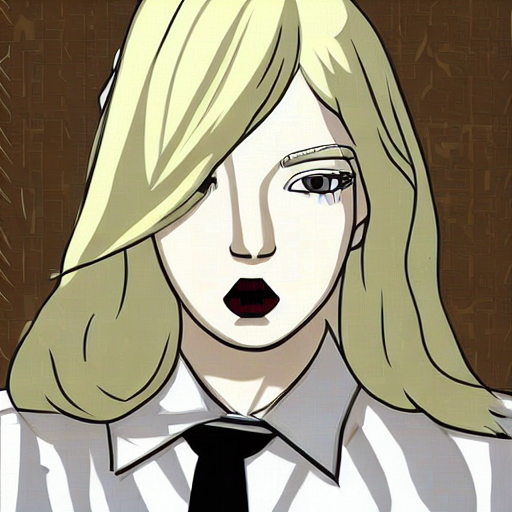}\hfill
\makebox[\w][c]{Clean}
\makebox[\w][c]{PGD}
\makebox[\w][c]{UAP}\hfill
\label{fig:pgd_uap}
\end{minipage}
\vspace{-1.0em}
\end{table}
}

\newcommand{\tableComparison}{
\begin{table*}[t]
\centering
\caption{Quantitative comparison. Latency is measured on 512$\times$512 image. Comparisons of other metrics are shown in Appendix.}
\label{tab:comparison}
\vspace{-0.5em}
\setlength\tabcolsep{13pt}
\begin{tabular}{l|c|cccc}
\hline
\multirow{2}{*}{Method}    & Latency & Object & Face  & Painting & Cartoon \\ 
\cline{2-6}
& CPU / GPU & \multicolumn{4}{c}{Invisibility: DISTS ($\downarrow$) / Efficacy: FID ($\uparrow$)} \\
\hline\hline
AdvDM~\cite{liang2023adversarial}     & 1210s / \hspace{1ex}35s & 0.197 / \underline{220.0}	& 0.173 / 303.8	& 0.153 / \textbf{357.6} & 0.271 / 212.5 \\
PhotoGuard~\cite{salman2023raising}   & \hspace{1ex}\underline{370s} / \hspace{2ex}\underline{7s}  & 0.203 / \textbf{223.0} & 0.189 / \underline{308.7} & \textbf{0.107} / 350.9 & 0.209 / 219.1 \\
Anti-DB~\cite{van2023anti}      & 7278s / 225s & 0.239 / 214.4 & 0.162 / 301.4 & 0.114 / 347.7 & 0.294 / \textbf{225.4} \\
Mist~\cite{liang2023mist}         & 1440s / \hspace{1ex}40s  & \underline{0.185} / 217.2 & \underline{0.154} / 307.5 & 0.129 / \underline{357.0} & 0.223 / 223.7 \\
Impasto~\cite{ahn2024imperceptible}      & \hspace{1ex}830s / \hspace{1ex}19s & 0.201 / 213.8 & 0.198 / 298.4 & 0.123 / 352.4 & \underline{0.207} / 215.5 \\
SDST~\cite{xue2023toward}          & 1410s / \hspace{1ex}24s & 0.242 / 219.2 & 0.244 / 302.1 & 0.152 / 354.1 & 0.237 / \underline{222.7} \\
\hline
\ours         & \textbf{\hspace{2.3ex}2.9s} / \textbf{0.04s} & \textbf{0.155} / \textbf{223.0} & \textbf{0.149} / \textbf{308.9} & \underline{0.110} / 356.1 & \textbf{0.186} / 220.3 \\
\hline
\end{tabular}
\end{table*}
}

\newcommand{\tableBlackbox}{
\begin{table*}[t]
\centering
\caption{Analysis of black-box protection scenarios (unknown diffusion models and personalization methods).}
\label{tab:blackbox}
\vspace{-0.5em}
\setlength\tabcolsep{11pt}
\begin{tabular}{l|l|c|ccccc}
\hline
\multirow{2}{*}{Domain} & \multirow{2}{*}{Method} & \multirow{2}{*}{Invisibility} & \multicolumn{2}{c}{Unknown Model} & \multicolumn{2}{c}{Unknown Personalization} \\
\cline{4-7}
& & & $\rightarrow$ SD-v2.1~\cite{rombach2022high} & $\rightarrow$ SD-XL~\cite{podell2023sdxl} & TI~\cite{gal2022image} & DreamStyler~\cite{ahn2023dreamstyler} \\
\hline\hline
\multirow{2}{*}{Subject} & PhotoGuard & 0.203 & 176.6 & 190.9 & 290.2 & 224.9 \\
& \ours & \textbf{0.155} & \textbf{177.5} & \textbf{218.4} & \textbf{305.6} & \textbf{231.3} \\
\hline
\multirow{2}{*}{Cartoon} & PhotoGuard & 0.209 & 179.3 & 195.1 & \textbf{306.4} & 193.7 \\
& \ours & \textbf{0.186} & \textbf{188.3} & \textbf{207.3} & 305.7 & \textbf{209.9} \\
\hline
\end{tabular}
\vspace{-1em}
\end{table*}
}

\newcommand{\tableRobustness}{
\begin{table*}[t]
\centering
\begin{minipage}{.3\linewidth}
\centering
\caption{Ablation study.}
\label{tab:ablation}
\vspace{-0.5em}
\bgroup
\setlength\tabcolsep{10pt}
\def\arraystretch{0.92}
\begin{tabular}{l|c}
\hline
Configuration & FID ($\uparrow$) \\
\hline\hline
PhotoGuard & 227.6 \\
UAP~\cite{moosavi2017universal} & 207.6 \\
\hline
MoP (w/o $\mathcal{A}$) & 214.5 \\
MoP & 225.9 \\
\hline
+ MLP Loss & 234.6 \\
+ Adapt. Target & 238.8 \\
\hline
\end{tabular}
\egroup
\end{minipage}\hfill
\begin{minipage}{.68\linewidth}
\centering
\bgroup
\def\arraystretch{1.075}
\setlength\tabcolsep{8pt}
\caption{Analysis on protection robustness scenarios.}
\vspace{-0.5em}
\label{tab:robustness}
\begin{tabular}{l|l|ccccc}
\hline
\multirow{2}{*}{Domain} & \multirow{2}{*}{Method} & \multirow{2}{*}{Invisibility} & \multicolumn{2}{c}{Countermeasure} & \multirow{2}{*}{Arbitrary Size} \\
\cline{4-5}
& & & Noise & JPEG \\
\hline\hline
\multirow{2}{*}{Subject} & PhotoGuard & 0.203 & 193.3 & \textbf{193.2} & 193.5 \\
& \ours & \textbf{0.155} & \textbf{214.4} & 191.3 & \textbf{219.1} \\
\hline
\multirow{2}{*}{Cartoon} & PhotoGuard & 0.209 & \textbf{192.9} & 193.1 & 190.7 \\
& \ours & \textbf{0.186} & 191.8 & \textbf{199.9} & \textbf{204.0}  \\
\hline
\end{tabular}
\egroup
\end{minipage}\hfill
\vspace{-0.5em}
\end{table*}
}

\newcommand{\tableTargetImage}{
\begin{table}[t]
\centering
\caption{Quantitative report on the effect of adaptive targeted protection shown in \fref{fig:analysis_target}.}
\label{tab:appendix_target_image}
\vspace{-0.5em}
\begin{tabular}{c|cccc}
\hline
\multirow{2}{*}{Target Image} & \multicolumn{4}{c}{Inference Domains} \\
\cline{2-5}
& Object & Face & Painting & Cartoon \\
\hline\hline
Low rep.   & 200.1	& \textbf{327.7}	& 347.5	& \textbf{237.3} \\
Mid rep.   & 207.2	& 297.9	& \underline{349.3}	& 234.6 \\
High rep.  & \underline{208.3}	& 270.7	& 348.5	& 211.8 \\
\hline
Adaptive target  & \textbf{208.5}	& \underline{320.2}	& \textbf{349.4}	& \underline{235.4} \\
\hline
\end{tabular}
\vspace{-1em}
\end{table}
}

\newcommand{\tableSupplSubject}{
\begin{table*}[t]
\centering
\caption{Additional quantitative comparison on the Subject domain.}
\label{tab:appendix_comp_subject}
\setlength\tabcolsep{10pt}
\vspace{-0.5em}
\begin{tabular}{l|ccc|ccc}
\hline
\multirow{2}{*}{Domain: Subject} & \multicolumn{3}{c|}{Invisibility} &  \multicolumn{3}{c}{Protection Efficacy} \\ 
\cline{2-7}
&	DISTS ($\downarrow$) &	$\text{LPIPS}_{\text{VGG}}$ ($\downarrow$)	& AHIQ ($\uparrow$) &	FID ($\uparrow$) &	TOPIQ-NR ($\downarrow$) & QAlign ($\downarrow$) \\
\hline\hline
AdvDM~\cite{liang2023adversarial}      &   0.197 &	0.362 &	0.540 &	\underline{220.0} &	\underline{0.458} &	\textbf{2.139} \\
PhotoGuard~\cite{salman2023raising}	   &   0.203 &	0.347 &	0.575 &	\textbf{223.0} &	0.506 &	2.396 \\
Mist~\cite{liang2023mist}	           &   \underline{0.185} &	\underline{0.322} &	0.578 &	217.2 &	0.523 &	2.328 \\
SDST~\cite{xue2023toward}              &   0.242 &	0.402 &	0.575 &	219.2 &	0.542 &	2.442 \\
Anti-DB~\cite{van2023anti}	           &   0.239 &	0.413 &	0.510 &	214.4 &	\textbf{0.439} &	\underline{2.148} \\
Impasto~\cite{ahn2024imperceptible}	   &   0.201 &	0.378 &	\textbf{0.588} &	213.8 &	0.473 &	2.183 \\
\hline
\ours	                               &   \textbf{0.155} &	\textbf{0.258} &	\underline{0.583} &	\textbf{223.0} &	0.507 &	2.364 \\
\hline
\end{tabular}
\end{table*}
}

\newcommand{\tableSupplFace}{
\begin{table*}[t]
\centering
\caption{Additional quantitative comparison on the Face domain.}
\label{tab:appendix_comp_face}
\setlength\tabcolsep{10pt}
\vspace{-0.5em}
\begin{tabular}{l|ccc|ccc}
\hline
\multirow{2}{*}{Domain: Face} & \multicolumn{3}{c|}{Invisibility} &  \multicolumn{3}{c}{Protection Efficacy} \\ 
\cline{2-7}
&	DISTS ($\downarrow$) &	$\text{LPIPS}_{\text{VGG}}$ ($\downarrow$)	& AHIQ ($\uparrow$) &	FID ($\uparrow$) &	TOPIQ-NR ($\downarrow$) & QAlign ($\downarrow$) \\
\hline\hline
AdvDM~\cite{liang2023adversarial}      &   0.173 &	0.338	& 0.528	& 303.8	& \textbf{0.466}	 & \underline{2.493} \\
PhotoGuard~\cite{salman2023raising}	   &   0.189 &	0.357	& 0.557	& \underline{308.7}	& 0.630	 & 3.113 \\
Mist~\cite{liang2023mist}	           &   \underline{0.154} &	0.310	& 0.560	& 307.5	& 0.667	 & 3.131 \\
SDST~\cite{xue2023toward}              &   0.244 &	0.431	& 0.561	& 302.1	& 0.629	 & 3.019 \\
Anti-DB~\cite{van2023anti}	           &   0.162 &	\underline{0.309}	& 0.543	& 301.4	& \underline{0.484}	 & \textbf{2.411} \\
Impasto~\cite{ahn2024imperceptible}	   &   0.198 &	0.388	& \underline{0.562}	& 298.4	& 0.565	 & 2.939 \\
\hline
\ours	                               &   \textbf{0.149} &	\textbf{0.280}	& \textbf{0.566}	& \textbf{308.9}	& 0.558	 & 2.851 \\
\hline
\end{tabular}
\end{table*}
}

\newcommand{\tableSupplPainting}{
\begin{table*}[t]
\centering
\caption{Additional quantitative comparison on the Painting domain.}
\label{tab:appendix_comp_painting}
\setlength\tabcolsep{10pt}
\vspace{-0.5em}
\begin{tabular}{l|ccc|ccc}
\hline
\multirow{2}{*}{Domain: Painting} & \multicolumn{3}{c|}{Invisibility} &  \multicolumn{3}{c}{Protection Efficacy} \\ 
\cline{2-7}
&	DISTS ($\downarrow$) &	$\text{LPIPS}_{\text{VGG}}$ ($\downarrow$)	& AHIQ ($\uparrow$) &	FID ($\uparrow$) &	TOPIQ-NR ($\downarrow$) & QAlign ($\downarrow$) \\
\hline\hline
AdvDM~\cite{liang2023adversarial}      &   0.153	& 0.267	& 0.572	& \textbf{357.6}	& \textbf{0.438}	& \underline{2.889} \\
PhotoGuard~\cite{salman2023raising}	   &   \textbf{0.107}	& 0.189	& \textbf{0.579}	& 350.9	& 0.613	& 3.228 \\
Mist~\cite{liang2023mist}	           &   0.129	& 0.219	& \textbf{0.579}	& \underline{357.0}	& 0.615	& 3.218 \\
SDST~\cite{xue2023toward}              &   0.152	& 0.258	& 0.583	& 354.1	& 0.637	& 3.288 \\
Anti-DB~\cite{van2023anti}	           &   0.114	& \underline{0.184}	& \textbf{0.579}	& 347.7	& \underline{0.452}	& \textbf{2.860} \\
Impasto~\cite{ahn2024imperceptible}	   &   0.123	& 0.221	& \textbf{0.579}	& 352.4	& 0.564	& 3.141 \\
\hline
\ours	                               &   \underline{0.110}	& \textbf{0.180}	& \underline{0.577}	& 356.1	& 0.506	& 2.987 \\
\hline
\end{tabular}
\end{table*}
}

\newcommand{\tableSupplWebtoon}{
\begin{table*}[t]
\centering
\caption{Additional quantitative comparison on the Cartoon domain.}
\label{tab:appendix_comp_webtoon}
\setlength\tabcolsep{10pt}
\vspace{-0.5em}
\begin{tabular}{l|ccc|ccc}
\hline
\multirow{2}{*}{Domain: Cartoon} & \multicolumn{3}{c|}{Invisibility} &  \multicolumn{3}{c}{Protection Efficacy} \\ 
\cline{2-7}
&	DISTS ($\downarrow$) &	$\text{LPIPS}_{\text{VGG}}$ ($\downarrow$)	& AHIQ ($\uparrow$) &	FID ($\uparrow$) &	TOPIQ-NR ($\downarrow$) & QAlign ($\downarrow$) \\
\hline\hline
AdvDM~\cite{liang2023adversarial}      &   0.271	& 0.440	& 0.445	& 212.5	& \textbf{0.579}	 & \underline{2.247} \\
PhotoGuard~\cite{salman2023raising}	   &   0.209	& \underline{0.348}	& 0.587	& 219.1	& 0.683	 & 2.618 \\
Mist~\cite{liang2023mist}	           &   0.223	& 0.365	& 0.585	& \underline{223.7}	& 0.681	 & 2.645 \\
SDST~\cite{xue2023toward}              &   0.237	& 0.398	& \underline{0.589}	& 222.7	& 0.712	 & 2.755 \\
Anti-DB~\cite{van2023anti}	           &   0.294	& 0.468	& 0.445	& \textbf{225.4}	& \textbf{0.429}	 & \textbf{2.166} \\
Impasto~\cite{ahn2024imperceptible}	   &   \underline{0.207}	& 0.376	& \textbf{0.602}	& 215.5	& 0.590	 & 2.294 \\
\hline
\ours	                               &   \textbf{0.186}	& \textbf{0.301}	& 0.585	& 220.3	& 0.656	 & 2.530 \\
\hline
\end{tabular}
\end{table*}
}

\newcommand{\tableComparisonSameBudget}{
\begin{table*}[t]
\centering
\caption{Quantitative comparison when fix the perturbation strength ($\eta$) to eight for all the protection frameworks.}
\label{tab:comparison_same_budget}
\vspace{-0.5em}
\setlength\tabcolsep{13pt}
\begin{tabular}{l|c|cccc}
\hline
\multirow{2}{*}{Method ($\eta=8$)}    & Latency & Object & Face  & Painting & Cartoon \\ 
\cline{2-6}
& CPU / GPU & \multicolumn{4}{c}{Invisibility: DISTS ($\downarrow$) / Efficacy: FID ($\uparrow$)} \\
\hline\hline
AdvDM~\cite{liang2023adversarial}   & 1210s / \hspace{1ex}35s & \underline{0.129} / 199.5	& \underline{0.152} / 303.7	& \underline{0.117} / 346.8 & 0.194 / 196.5 \\
PhotoGuard~\cite{salman2023raising}	& \hspace{1ex}\underline{370s} / \hspace{2ex}\underline{7s}  & 0.184 / \underline{211.9}	& 0.258 / \textbf{371.1}	& 0.165 / \underline{377.4} & 0.221 / \textbf{227.6} \\
Anti-DB~\cite{van2023anti} 	        & 7278s / 225s & 0.164 / 197.4	& 0.193 / 317.4	& 0.143 / 343.9 & 0.231 / 184.6 \\
Mist~\cite{liang2023mist}	        & 1440s / \hspace{1ex}40s & 0.185 / \textbf{217.2}	& 0.259 / \underline{365.8}	& 0.166 / \textbf{386.2} & 0.223 / \underline{223.7} \\
Impasto~\cite{ahn2024imperceptible} & \hspace{1ex}830s / \hspace{1ex}19s & 0.131 / 188.9	& 0.198 / 298.4	& 0.123 / 352.4 & \textbf{0.179} / 201.1 \\
SDST~\cite{xue2023toward} 	        & 1410s / \hspace{1ex}24s & 0.170 / 198.0	& 0.244 / 302.1	& 0.152 / 354.1 & 0.199 / 196.7 \\
\hline
\ours	                            & \textbf{\hspace{2.3ex}2.9s} / \textbf{0.04s} & \textbf{0.097} / 200.4	& \textbf{0.149} / 308.9	& \textbf{0.048} / 348.0 & \underline{0.186} / 220.3 \\
\hline
\end{tabular}
\vspace{-1em}
\end{table*}
}

%% file: sections/algorithms.tex
\newcommand{\algTrain}{
\begin{algorithm*}[t]
\SetNoFillComment
\DontPrintSemicolon
\caption{Training stage of \ours\ (Single-batch example)}
\label{alg:train}
\KwIn{Training dataset $\mathcal{X_D}$, target images $\{\mathbf{y}^l, \mathbf{y}^m, \mathbf{y}^h\}$, \# protection perturbations $K$, \# VAE encoder layers $L$, Perturbation budget (strength) $\eta$, Training steps $N$}
\KwOut{Assignment function $\mathcal{A}$, mixture-of-perturbation $\{\delta_g, \Delta\}$}
\tcc{Compute assignment function prior to train MoP}
$\mathcal{A} \gets \texttt{K-means++}(\mathcal{E}(\mathcal{X_D}), \, K)$ \;
\BlankLine
\tcc{MoP training}
Initialize $\{\delta_g,\, \Delta\} \gets 0$ \;

$\mathbf{z}^l_y, \, \mathcal{F}^l_y \gets \mathcal{E}(\mathbf{y}^l, \, L)$; \,\,
$\mathbf{z}^m_y, \, \mathcal{F}^m_y \gets \mathcal{E}(\mathbf{y}^m, \, L)$; \,\,
$\mathbf{z}^h_y, \, \mathcal{F}^h_y \gets \mathcal{E}(\mathbf{y}^h, \, L)$ \,\, \tcp{Prepare target images}
\For{$n \gets 1$ \KwTo $N$}{
    $\mathbf{x} \gets \texttt{sample-batch}(\mathcal{X}_{D})$; \,\, $\mathbf{z} \gets \mathcal{E}(\mathbf{x})$ \;
    $k \gets \mathcal{A}(\mathbf{z})$; \,\, $t \gets \argmin_{i \in {\{l, m, h\}}} ||\mathcal{H}(\mathbf{z}) - \mathcal{H}(\mathbf{z}_y^i)||_1$ \,\, \tcp{Refer to \eref{eq:mop} and \ref{eq:atp}}

    \BlankLine
    $\mathbf{x} \gets \mathbf{x} + \delta_g^t + \Delta_k^t$ \;
    $\mathbf{z}, \, \mathcal{F} \gets \mathcal{E}(\mathbf{x}, \, L)$ \;

    \BlankLine
    $\mathcal{L}_T = || \mathbf{z} - \mathbf{z}^t_y ||^2_2 + \frac{\lambda}{L} \sum_{l=1}^L || \mathcal{F}^{t, l} - \mathcal{F}^{t, l}_y ||^2_2$ \,\, \tcp{Multi-layer protection loss (\eref{eq:mlp_loss})}
    $\{\delta^t_g, \, \Delta^t_k\} \gets \texttt{Adam}(\nabla_{\{\delta^t_g, \Delta^t_k\}}\mathcal{L}_T)$ \;
    $\{\delta^t_g, \, \Delta^t_k\} \gets \texttt{Clamp}(\{\delta^t_g, \, \Delta^t_k\}, \, -\frac{\eta}{2}, \frac{\eta}{2})$ \;

}
\end{algorithm*}
}

\newcommand{\algInference}{
\begin{algorithm*}[t]
\SetNoFillComment
\DontPrintSemicolon
\caption{Inference stage of \ours}
\label{alg:inference}
\vspace{0.1em}
\KwIn{Image $\mathbf{x}$}
\KwOut{Protected image $\hat{\mathbf{x}}$}
$H, \, W \gets \texttt{size}(\mathbf{x})$ \;
$\mathbf{z} \gets \mathcal{E}(\texttt{resize}(\mathbf{x}; \, (512, 512)))$ \;
\BlankLine

\tcc{MoP with adaptive targeted protection}
$k \gets \mathcal{A}(\mathbf{z})$ \;
$t \gets \argmin_{i \in {\{l, m, h\}}} ||\mathcal{H}(\mathbf{z}) - \mathcal{H}(\mathbf{z}_y^i)||_1$ \,\, \tcp{Utilize pre-computed target entropy}
$\delta \gets \texttt{resize}(\delta^t_g+\Delta^t_k; \, (H, W))$ \,\, \tcp{Match MoP's resolution to input image}
\BlankLine

\tcc{Adaptive protection strength}
$\mathbf{M} \gets \texttt{LPIPS}(\mathbf{x}, \, \mathbf{x} + \delta)$ \,\, \tcp{Use surrogate protected image}
$\mathbf{M}' \gets \mathcal{S}(1 - \mathbf{M})$ \,\, \tcp{Refer to \eref{eq:scaling}}
$\hat{\mathbf{x}} = \mathbf{x} + \mathbf{M}' * \delta$
\vspace{0.2em}
\end{algorithm*}
}

%% file: sections/0_abstract.tex
\begin{abstract}

Recent advancements in diffusion models revolutionize image generation but pose risks of misuse, such as replicating artworks or generating deepfakes. Existing image protection methods, though effective, struggle to balance protection efficacy, invisibility, and latency, thus limiting practical use. We introduce perturbation pre-training to reduce latency and propose a mixture-of-perturbations approach that dynamically adapts to input images to minimize performance degradation. Our novel training strategy computes protection loss across multiple VAE feature spaces, while adaptive targeted protection at inference enhances robustness and invisibility. Experiments show comparable protection performance with improved invisibility and drastically reduced inference time.
The code and demo are available at \url{https://webtoon.github.io/impasto}
\end{abstract}

%% file: sections/1_introduction.tex
\section{Introduction}
Large-scale diffusion models~\cite{song2020score,ho2020denoising,rombach2022high,podell2023sdxl} have achieved remarkable success in the realm of image synthesis task, revolutionizing the way we create and manipulate digital imagery~\cite{zhang2023adding,kim2023diffblender,huang2023composer,meng2021sdedit,mou2023t2i}.
By adopting recently emerged personalization techniques, it is now possible to develop diffusion models that can generate images in one's unique styles~\cite{ruiz2023dreambooth,gal2022image,kumari2023multi,sohn2023styledrop,ahn2023dreamstyler}.
However, the very capabilities that make these models so valuable also pose significant risks.
Diffusion models can allow malicious users to replicate an individual's artwork without consent, easily stealing their creative output~\cite{shan2023glaze}.
Furthermore, diffusion models are exceedingly adept at mimicking faces and objects~\cite{ruiz2023dreambooth,kumari2023multi}.
This proficiency is not limited to benign applications but can be extended to the making of deepfakes or fake news.
The implications for societal harm are profound, as these technologies could be used to undermine public trust, distort political discourse, and violate personal privacy.

Current efforts to protect images against diffusion models have evolved through adversarial perturbation~\cite{goodfellow2014explaining}.
By introducing perturbations to a given image, they render the protected image resistant to diffusion-based personalization methods, causing the diffusion model to generate distorted outputs.
Existing methods primarily rely on two approaches: minimizing the distance between the latent features of the protected and a pre-defined target image within a VAE encoder (\eg PhotoGuard~\cite{salman2023raising} and Glaze~\cite{shan2023glaze}), or fooling the noise prediction network (\eg AdvDM~\cite{liang2023adversarial} and Anti-DB~\cite{van2023anti}).
Despite numerous advancements, all current frameworks fundamentally start from an approach that optimizes perturbations during inference~\cite{madry2017towards} when protection is requested, casting doubt on the practicality of these methods to be used in many real applications.

For a protection framework to become a practical solution, the following conditions should be satisfied:
\textbf{(1) Protection Efficacy}: Obviously, image protection method should effectively counteract against diffusion-based personalization.
Many recent studies have aimed to maximize this aspect by analyzing the internal dynamics of diffusion models~\cite{xue2023toward, zheng2023understanding}.
\textbf{(2) Invisibility}: Although some studies strive to be as invisible as possible, adversarial perturbations inevitably leave visible traces on images.
However, for commercial services, minimizing these traces is crucial.
Moreover, as analyzed by \citet{ahn2024imperceptible}, the task of preventing mimicry tends to leave more noticeable traces, particularly in images with flat textures  (\eg cartoon or illustration).
Despite its importance, only a handful of studies highlight this issue~\cite{ahn2024imperceptible}.
\textbf{(3) Latency}: All existing methods update perturbations during the inference phase through iterative optimization.
However, this process is inherently time-consuming; \eg protecting a $512^2$-px image can take up to 5-120 minutes on a CPU and 7-200 seconds even on a high-end A100 GPU.
The extensive time required by current protection methods poses a significant barrier that prevents ordinary users from utilizing image protection, thereby still exposing them to the risk of misuse of generative models.
Consequently, in order to democratize this technology, it is crucial to develop protection framework that can operate effectively on less powerful devices.
Unfortunately, despite extensive progresses in the image protection task, there is scant focus on reducing latency.

In this work, we propose a novel protection framework, \ours, which meets all the requirements, with a particular focus on latency.
\ours\ diverges from prevalent approaches by adopting a pre-trained perturbation.
Although universal adversarial perturbation (UAP)~\cite{moosavi2017universal} is proposed in adversarial attack, we observed that it significantly reduces protection efficacy for our task.
To counteract this, we introduce a \textbf{mixture-of-perturbations (MoP)}.
Unlike UAP, which relies on a single perturbation, MoP prepares multiple perturbations and selects one of them based on the input image's latent code.
We also propose a \textbf{multi-layer protection (MLP) loss} that utilizes the intermediate features when calculating the protection loss.
It intensifies the distortion when the protected images go through fine-tuning by personalization methods, thus enhancing the protection efficacy without any additional cost at inference.

For the inference, we introduce an \textbf{adaptive targeted protection} to enhance protection efficacy with minimal computational overhead.
Our observations suggest that in targeted protection, which minimizes the distance between the input image’s latent code and that of a (pre-defined) target, selecting an appropriate target image significantly influences protection performance.
Taking this into account, we adaptively determine the best target image based on the input image's latent code and then apply the perturbation trained for that specific target.
To enhance invisibility, we propose an \textbf{adaptive protection strength} approach using the LPIPS~\cite{zhao2023unlearnable} distance.
Unlike \citet{ahn2024imperceptible} that leverages multiple just-noticeable difference (JND) maps, our approach does not require ad-hoc modules nor does it significantly increase computational load; it only involves an additional forward pass of LPIPS (i.e. AlexNet~\cite{krizhevsky2012imagenet}).

We conduct extensive experiments to verify the effectiveness and efficiency of our protection method.
To simulate various use-cases, we test across diverse domains--- natural images, faces, paintings, and cartoons--- under different personalization and countermeasure scenarios.
\ours\ achieves similar protection efficacy to other methods but \textit{at nearly zero cost}, with a \textbf{200 to 3500$\times$ speedup} (\fref{fig:teaser_latency}).
It also shows an improvement in invisibility compared to most methods (\fref{fig:teaser_tradeoff}), proving its suitability for practical applications.
Our main contributions are:

\begin{itemize}[leftmargin=2em]
    \item We propose \ours, which achieves real-time protection against diffusion models. Our work is the first to address the critical issue of latency in this task.
    \item \ours\ integrates perturbations pre-training  with adaptive inference schemes, meeting all requirements for a practical protection solution.
    \item We validate \ours\ in various scenarios, showing that despite its speed and invisibility, it retains protection efficacy, robustness, and generalization.
\end{itemize}

%% file: sections/2_background.tex
\section{Background}

\textbf{Diffusion Models.}
These have gained prominence due to their ability to generate high-quality images.
Latent Diffusion Model (LDM)~\cite{rombach2022high} is particularly highligted for its exceptional quality and efficiency.
Within LDM, a VAE encoder $\mathcal{E}$, transforms an input image $\mathbf{x}$ into a latent code $\mathbf{z} = \mathcal{E}(\mathbf{x})$.
This code is then reconstructed back into the image domain by a decoder $\mathcal{D}$; $\mathbf{x}' = \mathcal{D}(\mathbf{z}')$.
The diffusion process makes a latent code $\mathbf{z}'$ by incorporating external factors, $y$, such as textual prompts.
The training of LDM is driven by a denoising loss function at each timestep $t$ as:
\begin{equation}
\mathcal{L_{SD}} = \mathbb{E}_{\mathbf{z}\sim E(\mathbf{x}), y, \epsilon\sim N(0, 1), t} [||\epsilon - \epsilon_{\theta}(\mathbf{z}_t, t, c(y))||^2_2],
\label{eq:sd}
\end{equation}
where a denoising network $\epsilon_{\theta}$ restores the noised latent code $\mathbf{z}_t$ based on the timestep $t$ and a conditioning vector $c(y)$.
Recent advancements explore the personalization (or few-shot fine-tuning) of LDM with a few reference images through two primary approaches: textual inversion~\cite{gal2022image}, which utilizes the embedding space of CLIP~\cite{radford2021learning} while keeping the denoising network fixed, and model optimization~\cite{ruiz2023dreambooth}, which directly updates the denoising network.

\smallskip
\noindent\textbf{Protection Against Diffusion-based Mimicry.}
Current protection frameworks apply adversarial perturbations $\delta$ to image $\mathbf{x}$, producing a protected image $\hat{\mathbf{x}}=\mathbf{x} + \delta$ using projected gradient descent (PGD)~\cite{madry2017towards}.
With a protection loss $\mathcal{L_P}$, current methods obtain the protected image of $i$-th optimization step by a signed gradient ascent with step function $\text{sgn}$ and step length $\alpha$ as given by
\begin{equation}
\mathbf{x}^{(i)} = \Pi_{\mathcal{N}_{\eta}(\mathbf{x})} \left[ \mathbf{x}^{(i-1)} + \alpha \text{sgn} ( \nabla_{\mathbf{x}^{(i)}}\mathcal{L_{P}}(\mathbf{x}^{(i-1)}) \right],
\end{equation}
where $\Pi_{\mathcal{N}{\eta}(\mathbf{x})}$ projects onto the $L_{\infty}$ neighborhood of $\mathbf{x}$ with a radius of $\eta$.
This iterative process is repeated for $N$ steps until $\hat{\mathbf{x}} = \mathbf{x}^{(N)}$ is achieved (\fref{fig:model_overview}a).

For the objective function, semantic or texture losses are mostly used~\cite{liang2023mist}.
The semantic loss, $\mathcal{L_S}(\mathbf{x}) = \mathcal{L_{SD}}(\mathbf{x})$, is designed to disrupt the denoising process of LDM, misleading it to generate samples that deviate from the original images.
AdvDM~\cite{liang2023adversarial} adopts this loss and Anti-DB~\cite{van2023anti} is improved to be more robust against DreamBooth~\cite{ruiz2023dreambooth} by incorporating its update process within the optimization.
The texture loss, $\mathcal{L_T}(\mathbf{x}) = -|| \mathcal{E}(\mathbf{x}) - \mathcal{E}(\mathbf{y}) ||^2_2$, aims to pushing the latent code of $\mathbf{x}$ towards the target image's latent code.
Glaze~\cite{shan2023glaze} and PhotoGuard~\cite{salman2023raising} belong to this category.
Upon these objective functions, We can now define a universal loss as $\mathcal{L_P} = \lambda_{\mathcal{S}}\mathcal{L}_S + \lambda_{\mathcal{T}}\mathcal{L}_T$, where $\{\lambda_{\mathcal{S}}, \lambda_{\mathcal{T}}\} \geq 0$ are balancing factors.
Mist~\cite{liang2023mist} utilizes this universal loss function to effectively capture the characteristics of both objectives. Diff-Protect~\cite{xue2023toward}, building on Mist, introduces a score distillation trick to achieve more efficient protection. Impasto~\cite{ahn2024imperceptible} incorporates perceptual-oriented components  designed for imperceptible protection.

Since all the current protection frameworks leverage iterative optimization during inference, they can yield image-specific and high-performing perturbations~\cite{salman2023raising,liang2023mist,shan2023glaze,liang2023adversarial,ahn2024imperceptible,van2023anti,xue2023toward,zheng2023understanding,xu2024perturbing,wan2024prompt,liu2024metacloak}.
Despite this advantage, a significant drawback is the substantial time required to operate protection.
For instance, to protect a $512^2$-px image, even the fastest model (\eg PhotoGuard~\cite{salman2023raising}) requires around 7 seconds on an A100 GPU and 350 seconds on a CPU.
Consequently, we explore alternatives to this iterative process to enhance efficiency without compromising the protection.

\figModelOverview{}

\smallskip
\noindent\textbf{Universal Adversarial Perturbation (UAP).}
This concept is introduced in adversarial attacks, through a data-driven approach to find an \textit{image-agnostic} perturbation from the training dataset~\cite{moosavi2017universal}.
In our task, we can adapt UAP as in \fref{fig:model_overview}b.
Given the training dataset $\mathcal{X_D}$, we employ a protection loss $\mathcal{L_P}$ to derive an universal perturbation $\delta$, constrained within $\eta$-ball to maintain perceptual invisibility.
\begin{equation}
\delta = \argmax_{||\delta||_\infty \leq \eta} \; \mathbb{E}_{\mathbf{x} \sim \mathcal{X_D}} \left[ \mathcal{L_P}(\mathbf{x} + \delta) \right].
\end{equation}

Upon finalizing the universal perturbation $\delta$ by the training, it can be applied to an image $\mathbf{x}$ to produce a protected image $\hat{\mathbf{x}} = \mathbf{x} + \delta$.
This is notably practical as it requires no computation at inference.
Nevertheless, we found that directly applying UAP to our task significantly compromises protection performance.
In \tref{table:pgd_uap}, we compare the protection efficacy of iterative optimization (via PGD) and UAP.
We use DISTS~\cite{ding2020image} to assess protected images' quality, while FID~\cite{heusel2017gans} evaluates generated image through mimicry by LoRA~\cite{hu2021lora} (a higher FID is better protection).
As also shown in the qualitative mimicry results by LoRA, UAP leads to a notable degradation in protection efficacy.
Hence, it is crucial to refine this pre-training approach, but with a minimal inference cost in a manner akin to UAP.

\tablePGDUAP{}

%% file: sections/3_method.tex
\section{Method}

\subsection{Perturbation Pre-Training}
\label{sec:method1}

\textbf{Mixture-of-Perturbation (MoP).}
We hypothesize that the underperformance of UAP is primarily due to two reasons: 1) a single perturbation has a limited capacity to deceive diffusion models, and 2) the image-agnostic nature of UAP fails to cover the diverse features (\eg texture or structure) of images.
To overcome these, we introduce a mixture-of-perturbations (MoP), consisting of multiple perturbations, $\Delta = \{\delta_1, \ldots, \delta_K\}$, where $K$ represents the number of perturbations.
For any given input image, MoP dynamically assigns and applies the appropriate perturbation to create a protected image (\fref{fig:model_overview}c).
Specifically, given an image $\mathbf{x}$, MoP first encodes it using a VAE encoder $\mathcal{E}$ to obtain the latent code $\mathbf{z}$.
This code is used to select the specific perturbation to be applied, as dictated by an assignment function $\mathcal{A}$.
Formally, the protected image $\hat{\mathbf{x}}$ is generated as:
\begin{equation}
\hat{\mathbf{x}} = \mathbf{x} + \delta_g + \Delta_k, \;\;\text{where} \;\; k = \mathcal{A}(\mathcal{E}(\mathbf{x})),
\label{eq:mop}
\end{equation}
where $\delta_g$ is a global perturbation.
We observed that adding $\delta_g$ slightly improves the performance.
In MoP, the assignment function $\mathcal{A}$ plays a crucial role by guiding the selection of perturbations based on an input image, ensuring that similar images are protected with the same perturbation.
This capability enables MoP to not only increase the capacity of UAP but also to offer a degree of image-specific protection, which is notably absent in UAP.
Without the assignment function, a naive perturbation averaging, \ie $\hat{\mathbf{x}} = \mathbf{x} + \delta_g + 1/K * \Sigma^K_{k=1}\Delta_k$, increases the capacity but fail to overcome the intrinsic limitations of image-agnostic, thus limiting performance benefits.
In contrast, our MoP implements a \textit{semi-image-specific} protection, positioning it between the image-agnostic and the image-specific precision of the iterative approach while removing the expensive inference cost, as demonstrated in the ablation study.

We adopt a simple assignment function $\mathcal{A}$ that can leverage the highly representative latent codes of the VAE encoder.
For our training dataset $\mathcal{X}_D$, we extract the latent codes $\mathcal{Z} = \{\mathbf{z}_1, \ldots, \mathbf{z}_D\}$ using the VAE encoder $\mathcal{E}$.
These codes are then clustered into $K$ predefined groups with K-means++~\cite{arthur2007k}.
Despite its simplicity, this process shows its robustness empirically, thus we adopt it for our framework.

\smallskip
\noindent\textbf{Multi-Layer Protection Loss.}
\ours\ employs a targeted protection approach based on the texture loss.
Conventionally, texture loss is computed upon the VAE latent code $\mathbf{z}$, \ie $\mathcal{L}_T(\mathbf{x}) = -|| \mathbf{z} - \mathbf{z}_y ||^2_2$, where $\mathbf{z}_y$ is the latent code of the target image.
However, we observed that often the $z$-space loss does not sufficiently push $\mathbf{z}$ towards $\mathbf{z}_y$, resulting in suboptimal protection performance.
To address this, we incorporate an auxiliary loss using multi-layer features extracted by the VAE encoder.
Specifically, given an input image, we extract the intermediate features $\mathcal{F} = \{\mathbf{f}^1, \ldots, \mathbf{f}^L\}$, where $L$ is the number of feature layers extracted.
Similarly, we obtain $\mathcal{F}_y$ for the target image.
Then, our proposed multi-layer protection loss function with the balanced factor $\lambda$ is computed as:
\begin{equation}
\mathcal{L}_T(\mathbf{x}) = -|| \mathbf{z} - \mathbf{z}_y ||^2_2 - \frac{\lambda}{L} \sum_{l=1}^L || \mathcal{F}^l - \mathcal{F}^l_y ||^2_2.
\label{eq:mlp_loss}
\end{equation}

The auxiliary MLP loss improves overall protection efficacy as this leverages the intermediate feature spaces beyond the $z$-space when pushing $\mathbf{z}$ toward $\mathbf{z}_y$.
In addition, since these modifications are implemented during the pre-training stage, it improves protection efficacy without extra computational expense at inference.

\smallskip
\noindent\textbf{Training.}
The assignment function $\mathcal{A}$ is first trained on $\mathcal{X_D}$ then MoP is updated on $\mathcal{X_D}$ using \eref{eq:mlp_loss} by Adam optimizer~\cite{kingma2014adam}.
All the perturbations ($\delta_g, \Delta$) are initialized at a resolution of $512\times512$.
When applied to input images of different resolutions, we resize the perturbation through bilinear interpolation.
Since two perturbations are used in MoP, $\delta_g, \Delta$ are constrained within a $(\eta/2)$-ball.

\subsection{Adaptive Inference}
\label{sec:method2}

\textbf{Adaptive Targeted Protection.}
In targeted protection, target image influences protection performance.
\citet{liang2023mist} noted that patterned image is effective, while \citet{zheng2023understanding} reported that the protection efficacy varies with pattern repetitions.
Though they provide valuable insights, they do not consider the relationship between the target and input images.
Instead, we investigate such relationship and propose an adaptive targeted protection approach.

\figTarget{}

In our analysis, we use two target images characterized by either low or high pattern repetition (shown in \suppl) and apply protection to input images with either simple or complex textures.
\fref{fig:target_observation} shows mimicry outputs by LoRA of these four protection scenarios.
Interestingly, our observations suggest a relationship between the pattern repetition of the target image and the texture complexity of an input image.
\citet{zheng2023understanding} noted that protection efficacy improves as pattern repetition increases but declines when it becomes too high.
We suspect that their results might be biased by fixing the input image, thereby keeping the texture factor constant.
In contrast, our findings indicate that tailoring the pattern image to match the texture complexity of an input image can improve protection efficacy.

Hence, we design \ours\ to adaptively select the target image based on an input image.
To this end, we prepare target images with varying pattern repetitions and train MoPs for each.
We utilize three pattern images---low, mid, and high---resulting in three MoPs: $\{\delta_g^l, \Delta^l\}$, $\{\delta_g^m, \Delta^m\}$, and $\{\delta_g^h, \Delta^h\}$.
We then extract the latent codes for each target image: $\mathbf{z}_y^l = \mathcal{E}(\mathbf{y}^l),\; \mathbf{z}_y^m = \mathcal{E}(\mathbf{y}^m),\; \mathbf{z}_y^h = \mathcal{E}(\mathbf{y}^h)$, where $\mathbf{y}^l, \mathbf{y}^m, \mathbf{y}^h$ are the target images with low, mid, and high pattern repetitions, respectively.
At inference, input image's latent code $\mathbf{z}$ is compared with $\{\mathbf{z}_y^l, \mathbf{z}_y^m, \mathbf{z}_y^h\}$ and the target image whose latent code is closest to $\mathbf{z}$ is selected.
To measure the similarity between the target images' pattern repetition and the input image, $L_1$ norm of entropy is used as a distance.
Overall, integration of MoP (in \eref{eq:mop}) and the adaptive targeted protection are as follows:
\begin{equation}
\hat{\mathbf{x}} = \mathbf{x} + \delta_g^t + \Delta_k^t, \;\; t = \argmin_{i \in {\{l, m, h\}}} ||\mathcal{H}(\mathbf{z}) - \mathcal{H}(\mathbf{z}_y^i)||_1,
\label{eq:atp}
\end{equation}
where $\mathcal{H}(\mathbf{z}) = -\sum_{z \in \mathbf{z}} p(z) \log p(z)$.
The major advantage of our adaptive targeted protection is its robust performance across various domains, which is critically important for practical use cases where the specific domain (or texture) of incoming images might be unknown.
If the range of potential domains is known beforehand, the performance difference might not be significant.
However, the flexibility to adapt to different input characteristics without prior domain knowledge enhances \ours's utility in diverse and unpredictable environments of real applications.

\figLPIPS{}

\tableComparison{}
\figQual{}

\smallskip
\noindent\textbf{Adaptive Protection Strength.}
According to the Weber–Fechner law, humans are more adept at detecting subtle changes in regions with simple textures than in complex ones.
Upon this, \citet{ahn2024imperceptible} adapted human visual system to the image protection task.
They utilized multiple perception maps and adjusted perturbation strength during optimization to effectively enhance invisibility.
However, the perception maps used by \citet{ahn2024imperceptible} are computed before the injection of perturbations, leading us to hypothesize that these may not perfectly align with the actual perceived perturbations intended for protection.
Moreover, their reliance on a combination of traditional perceptual map algorithms resulted in slow processing and performance limitations.
To address these challenges, we first apply MoP to create a \textit{surrogate protection image} and then generate a perceptual map, which helps resolve the identified issues.

Specifically, we generate a surrogate protection image $\hat{\mathbf{x}}$ through \eref{eq:atp}.
Subsequently, we create a spatial perceptual map using LPIPS~\cite{zhang2018unreasonable}, \ie $\mathbf{M} = \text{LPIPS}(\mathbf{x}, \hat{\mathbf{x}})$.
As shown in \fref{fig:lpips_map}, the perceptual map $\mathbf{M}$ is remarkably aligned with human cognitive perspectives.
Hence, we utilize this map to produce the final protected image as in below:
\begin{equation}
\hat{\mathbf{x}} = \mathbf{x} + \mathcal{S}(1 - \mathbf{M}) * (\delta^t_g + \Delta^t_k).
\label{eq:aps}
\end{equation}
Here, $\mathcal{S}(\cdot)$ is a scaling function (more details are in \suppl) and $\mathbf{M}$ serves as a distance map, necessitating an inversion step.
Since this process requires only the forward pass of the LPIPS backbone, the additional computational cost at inference is minimal, thus economically enhancing invisibility without significant overhead.
In \suppl, we summarize the perturbation pre-training and inference process in the form of an algorithm.

%% file: sections/4_experiments.tex
\section{Experiment}

\textbf{Implementation Details.}
We set the number of perturbations $K$ in MoP as four.
Other hyperparameters and details on training and inference are described in \suppl.

\smallskip
\noindent\textbf{Datasets.}
We utilize four domains when constructing both the training and benchmark datasets: object, face, painting, and cartoon.
These multiple scenarios allow us to analyze the models' protection performance more closely to real applications, as images from various domains can be encountered.
The dataset details are provided in \suppl.

\smallskip
\noindent\textbf{Baselines.}
We compare with existing diffusion-based image protection frameworks: AdvDM~\cite{liang2023adversarial}, PhotoGuard~\cite{salman2023raising}, Anti-DB~\cite{van2023anti}, Mist~\cite{liang2023mist}, Impasto~\cite{ahn2024imperceptible}, and Diff-Protect (SDST)~\cite{xue2023toward}.
Except ours, all other methods rely on iterative optimization, using texture or semantic losses.

\smallskip
\noindent\textbf{Evaluation.}
The protection efficacy is evaluated against LoRA~\cite{hu2021lora} on Stable Diffusion v1.5.
Latency is measured using an M1 Max CPU and an A100 GPU.
We employ DISTS~\cite{ding2020image} to evaluate invisibility, and FID~\cite{heusel2017gans} to quantify protection efficacy.
In \suppl, we provide additional metrics for a more comprehensive comparison.
Note that previous studies typically fix the protection strength across models for comparison, but we found that protection efficacy and invisibility vary significantly depending on the protection loss used. For instance, PhotoGuard~\cite{salman2023raising} and AdvDM~\cite{liang2023adversarial} exhibit different trends since they use disparate objectives. Moreover, evaluating protection efficacy only without invisibility leads to a narrow assessment that might overlook real-world needs. To address these, we adjust the protection strength to match the protection level across methods to ensure a fairer comparison and better reflection of practical requirements.
However, one might also be curious about how the performance appears when the protection strength is equalized across all methods. Hence, we provide evaluation with a fixed strength in \suppl.

\subsection{Model Comparison}
As shown in \fref{fig:teaser_latency} and \tref{tab:comparison}, \ours\ is ultra-fast; 125$\times$ faster on CPU and 175$\times$ faster on GPU compared to the second fastest model, PhotoGuard~\cite{salman2023raising}. Remarkably, \ours\ maintains consistent latency even as the input image size increases, achieving near real-time performance even for $2048^2$-px images. In contrast, all other frameworks show exponential increases in latency (\fref{fig:teaser_latency}), which poses a significant issue in the real world where many recent artworks are high-resolution. In addition, \ours\ occupies only 1.7GB of VRAM during inference, unlike other methods that require more than 8GB.
The unprecedented speed of our model makes it highly user-friendly.

\tableRobustness{}
\tableBlackbox{}

\ours\ also demonstrates a superior trade-off between protection efficacy and invisibility (\fref{fig:teaser_tradeoff}).
In comparisons where the protection strength is adjusted to yield similar protection efficacy (\tref{tab:comparison}), \ours\ consistently achieves the best invisibility across most protection domains, with the second-best performance in the painting domain. In our analysis, \ours\ shows the strong advantage in invisibility within areas that have flat textures. These features result in significant improvements in the object and cartoon domains. For example, \fref{fig:qual} shows that in the first row, the sky region is rendered nearly invisible by \ours, and the same applies to the cartoon in the second row.
As demonstrated in \suppl, even when we fix the protection strength of all methods to the same value, our method achieves a favorable balance in this trade-off.

\subsection{Model Analysis}
\label{sec:model_analysis}

\textbf{Ablation Study.}
In \tref{tab:ablation}, we depict the component analysis of \ours.
It is notable that UAP~\cite{moosavi2017universal} significantly deteriorates protection efficacy compared to the iterative optimization baseline, PhotoGuard~\cite{salman2023raising}.
On the other hand, when we introduce the proposed MoP, it recovers the lost performance.
However, if the assignment function $\mathcal{A}$ is not used, the improvement is limited, highlighting the importance of achieving a semi-image-specific nature.
As we discussed, the limitations of UAP arise not only from insufficient capacity but also from its image-agnostic nature. Simply increasing the number of perturbations (\ie MoP w/o $\mathcal{A}$) resolves the first limitation and improves performance over UAP. However, this still does not overcome the second one, which causes it to underperform compared to PhotoGuard. When we introduce the assignment function into MoP, it can dynamically account for image features. This allows MoP to match the performance of iterative optimization methods while remaining significantly faster.

When we attach the multi-layer protection (MLP) loss, it significantly increases the protection efficacy.
Similarly, adaptive targeted protection also enhances performance.
Overall, by incorporating novel modules in both the pre-training and inference stages, \ours\ can achieve better protection efficacy with a much faster inference.

\figTrainAnalysis{}
\figInferenceAnalysis{}

\smallskip
\noindent\textbf{Robustness.}
Here, we analyze the vulnerability of protected images to countermeasures (Table~\ref{tab:robustness}). We use PhotoGuard~\cite{salman2023raising} as our baseline due to its similarity to our approach. To assess robustness, we evaluate against countermeasures such as Gaussian noise and JPEG compression. Additionally, we account for a more realistic scenario where the input image can have arbitrary dimensions.
Most previous evaluations in image protection tasks assume a fixed image resolution, typically $512^2$, which represents an idealized scenario. In practice, image sizes and aspect ratios vary. Our model demonstrates comparable performance against these countermeasures, and notably, in the arbitrary size scenario, \ours\ performs even better. This suggests that simply using interpolation to adapt MoP to different sizes is both straightforward and effective.

\smallskip
\noindent\textbf{Black-Box Scenario.}
\tref{tab:blackbox} presents the model analysis in black-box scenarios. We examine two cases: the unknown model scenario, where we transfer protection methods to other Stable Diffusion (SD) backbones, specifically SD v2.1~\cite{rombach2022high} and SD-XL~\cite{podell2023sdxl}. We also assess image protection against unknown personalization by applying Textual Inversion (TI)~\cite{gal2022image} and DreamStyler~\cite{ahn2023dreamstyler} to replicate the image.
In these scenarios, \ours\ demonstrates superior invisibility compared to the baseline while achieving comparable or better performance in black-box settings.

\smallskip
\noindent\textbf{Mixture-of-Perturbation.}
In \fref{fig:analysis_domain}, we vary the training dataset $\mathcal{X_D}$ and measure the protection efficacy.
Interestingly, even when the training domain is limited, there is no significant gap in the protection performance.
We conjecture that MoP effectively handles unseen domains thanks to its adaptive perturbation selection mechanism.
However, using all domains is slightly better, so we train with this strategy.
Additionally, \fref{fig:analysis_k} compares \ours\ by changing $K$ and excluding the assignment function $\mathcal{A}$ in MoP.
Without the assignment function, performance improvement is limited, and in the full MoP case, it achieves substantially better efficacy with convergence at $K=4$.
In \fref{fig:analysis_group}, we visualize the representative images assigned to each perturbation when $K=4$.
The images are grouped by their certain distinguishing features (e.g., texture, scene).

\smallskip
\noindent\textbf{Adaptive Targeted Protection.}
\fref{fig:analysis_target} analyzes the relationship between protection efficacy and the target image. When a target image with low pattern repetition is used, the protection is effective in the Face and Cartoon domains but less so for Objects. In contrast, a high-repetition target image leads to strong performance in the Object domain.
The proposed adaptive targeted protection demonstrates near-optimal performance across all scenarios. Since it is difficult to predict which domain will be encountered in real-world use cases, relying on a single domain with a fixed target image is not practical. Therefore, incorporating adaptive protection into \ours\ is essential for creating a more robust protection solution in real-world applications.

\smallskip
\noindent\textbf{Adaptive Protection Strength.}
We analyze the impact of using the proposed adaptive protection strength in \fref{fig:analysis_lpips} by adjusting the perturbation budget.
Without this module, the trade-off between protection efficacy and invisibility is worse than that of the full model.
Note that when the budget is small, the perturbation is inherently minimal, so the difference could be marginal.
However, with stronger protection, the difference becomes significant.

\smallskip
\noindent\textbf{\ours\ + Iterative Optimization.} 
\ours\ is extremely fast, so users with sufficient computing resources might want to invest additional computation to achieve even more effective protection.
Considering this scenario, in \fref{fig:analysis_add_pgd}, we use the results of \ours\ as the initial perturbation and further refine them using PGD (\eg PhotoGuard~\cite{salman2023raising}).
Note that we display the baseline, PhotoGuard, with a slightly larger budget since applying PGD to \ours\ tends to decrease invisibility.
Hence, the baseline budget is adjusted to match the invisibility level of \ours\ + PGD.
Surprisingly, our method serves as a superior initial checkpoint for the iterative optimization techniques.
For example, the baseline, requires 100 steps to match our \textit{initial} results, whereas \ours\ + PGD converges in just 25 steps with a significantly higher protection efficacy.

%% file: sections/5_conclusion.tex
\section{Conclusion}
\label{sec:conclusion}
In this work, we propose \ours, which leverages pre-trained mixture-of-perturbations for low latency.
We also propose adaptive inference to compensate the loss of the protection efficacy and to improve the invisibility of the perturbation.
Our experiments demonstrate that \ours\ offers a more practical solution with comparable protection performance to existing methods, with improved invisibility and substantially reduced inference time.

\smallskip
\noindent\textbf{Limitations.}
\ours\ still produces some visible distortion and this is an unavoidable drawback when using adversarial perturbations. Future research should focus on finding new paradigms that can maximize the quality.


%% file: sections/6_appendix.tex
\section{Implementation Details}
\label{sec:appendix_implementation}
\textbf{Training.}
We set the number of perturbations \(K\) in MoP to four.
When calculating MLP loss in \eref{eq:mlp_loss}, we use intermediate features from [``down\_1", ``down\_2", ``down\_3", ``mid\_0"] layers in SD VAE, and set \(\lambda\) to \(3.5 \times 10^{-5}\).
We train \ours\ on a single A100 80GB GPU in an end-to-end manner except the assignment function $\mathcal{A}$.
We use a batch size of 16, using the Adam optimizer~\cite{kingma2014adam} for 40k steps with learning rate of 0.0002, and betas of (0.5, 0.99). When training, the required VRAM and compute time are around 50GB and 12 hours, respectively.
Note that since we update the perturbations with Adam instead of the direct optimization through PGD, we remove the minus sign in the training loss (\eref{eq:mlp_loss}) as: $\mathcal{L}_T(\mathbf{x}) = || \mathbf{z} - \mathbf{z}_y ||^2_2 + \frac{\lambda}{L} \sum_{l=1}^L || \mathcal{F}^l - \mathcal{F}^l_y ||^2_2.$
We utilize three patterned target images, each representing low, mid, and high pattern repetition, as illustrated in \fref{fig:target_example}.
To implement adaptive targeted protection, for each protection strength (budget) $\eta$, we obtain three MoP models (low, mid, and high), each corresponding to one of the target images.
The training algorithm is detailed in \aref{alg:train}; note that for simplicity, it is demonstrated with a batch size of one, but increasing the batch size is straightforward.

\smallskip
\noindent\textbf{Inference.}
When an input image with resolution other than $512^2$-px is given, we perform bilinear interpolation to the perturbations $(\delta_g, \Delta)$ to match the resolution with the input image's.
However, the SD VAE encoder receives downsampled image to a fixed $512^2$-px to prevent a significant increase in computational load during VAE encoding.
Although relative low-resolution is given to VAE, empirically, this shows robust performance to some extent.
For adaptive targeted protection, we pre-compute and cache the average entropy of the target images.
When an input image is given, we calculate its entropy and select the nearest target image based on the cached values.
Hence, adaptive targeted protection incurs minimal overhead since \ours\ already obtains $\mathbf{z}$ in the MoP assignment stage.
For the adaptive protection strength, we provide the surrogate protected image to LPIPS with its original resolution, since the LPIPS network (AlexNet) is significantly smaller than the VAE encoder thus it consumes negligible computational overhead.

When obtaining the final perceptual map in \eref{eq:aps}, we first apply min-max normalization to the spatial distance map calculated by LPIPS and then subtract it from one to reverse it.
Subsequently, we adjust scale of the perceptual map with scale function $\mathcal{S}(\cdot)$.
With a given perceptual map $\mathbf{M}$, which has resolution of  $H \times W$, we first initialize scaled perceptual map $\mathbf{M}'$ as one, i.e., $\mathbf{M}' = \mathbf{1}^{H \times W}$.
Then, we calculate the final perceptual map as follows:
\begin{equation}
\label{eq:scaling}
    \mathbf{M}'[\mathbf{M} < q_i] =
        \begin{cases}
        \beta^{i} \times \alpha & \text{if } i < c \\
        \beta^{i} & \text{if } i \geq c
        \end{cases},
\end{equation}
where $q_i = \text{decile}(\mathbf{M}, i)$. The function $\text{decile}(\mathbf{M}, i)$ computes the $i$-th decile value of $\mathbf{M}$ (from the highest value).
This scaling function is designed to perform stepwise scaling to the perceptual map.
For each region corresponding to a specific decile, we downscale by a factor of $\beta$.
Without this scaling, many regions would take very small perturbations, significantly reducing the overall protection efficacy.
For the areas up to the $c$-th decile, we additionally multiply by $\alpha$ to increase the perturbation intensity in the least noticeable regions.
This compensates for the overall perturbation magnitude lost in more noticeable regions.
In our work, we use $(\alpha, \beta, c) = (1.3, 0.91, 3)$.
The inference algorithm is detailed in \aref{alg:inference}.

\figTargetExample{}
\algTrain{}
\algInference{}

\section{Related Work}
Our protection method involves injecting adversarial perturbations into input images, a method extensively studied in the adversarial attack domain. In this section, we discuss two areas closely related to our motivation.

\smallskip
\noindent\textbf{Universal Adversarial Perturbation (UAP).}
\citet{moosavi2017universal} demonstrated that a single perturbation could be used to attack various images, in contrast to the traditional image-specific perturbations.
This approach is significantly more efficient in terms of computation time compared to per-instance adversarial attacks because it does not require iterative optimization for each individual image.
Subsequently, \citet{hayes2018learning} presented a method for generating UAPs using generative models, showing improved efficiency and effectiveness over previous methods.
Similarly, \citet{mopuri2018nag} utilized generative networks to create adversarial examples.
In addition, there have been various attempts to find UAPs tailored to the image recognition tasks~\cite{chaubey2020universal,liu2019universal,mopuri2018generalizable,mopuri2018ask}.
However, these methods are not aligned with and cannot be adapted to the image protection task.
Instead, in our study, we designed MoP with a focus on creating a protection framework that is both fast and maintains high performance. Our approach is specialized for image protection, differing from the general UAP-related methodologies introduced in adversarial attack field.

\tableTargetImage{}
\tableComparisonSameBudget{}

\smallskip
\noindent
\textbf{Invisible Adversarial Perturbation.}
Achieving high invisibility in adversarial attacks has garnered significant research interest. Several methods focus on restricting the regions of perturbations. Some approaches use the $L_0$ norm to generate sparse perturbations~\cite{modas2019sparsefool,croce2019sparse}, while others confine perturbations to small, salient areas to reduce visual distortions~\cite{dai2023saliency}.
Various studies have targeted specific elements such as low-frequency components~\cite{guo2018low,luo2022frequency} and have utilized advanced constraints like color components~\cite{zhao2020towards,shamsabadi2020colorfool} or quality assessments~\cite{wang2021demiguise}. Additionally, \citet{luo2018towards} explored techniques for creating invisible and robust adversarial examples based on the human visual system.
More recently, \citet{laidlaw2020perceptual} investigated perceptual adversarial robustness and adopted LPIPS~\cite{zhang2018unreasonable} in adversarial perturbation optimization.

Although invisibility has been highly emphasized in adversarial attacks, it is less explored in the image protection task.
Many methods enhance invisibility implicitly by minimizing the budget used, but this often leads to a trade-off with protection efficacy, resulting in decreased performance. \citet{ahn2024imperceptible} constrain perturbations based on the human visual system using various modules during the optimization process. While effective, they require many ad-hoc components.
In contrast, \ours\ uses LPIPS to create perceptual maps and apply masking, which is a simpler and more streamlined approach. This differs from \citet{zhang2018unreasonable}, who use LPIPS in the optimization process.

\section{Experimental Setups}
\label{sec:appendix_setups}

\textbf{Datasets.}
We utilize the \textit{Object}, \textit{Face}, \textit{Painting}, and \textit{Cartoon} domains in both the training and benchmark datasets.
When constructing the training dataset, we randomly sample 20k images from ImageNet~\cite{deng2009imagenet} for the Object domain, resizing all images to $512^2$.
No augmentation is applied. For the Face domain, we use 20k images randomly sampled from the FFHQ dataset~\cite{karras2019style}.
The images, originally $1024^2$ and face-aligned~\cite{kazemi2014one}, are all downscaled to $512^2$, without further augmentations.
The Painting dataset is created by randomly sampling 20k images from the WikiArt dataset~\cite{saleh2015large}, resizing them to $512^2$. We filter out images with resolutions outside the range of 512 to 2048-px.
For the Cartoon dataset, we use Webtoon artworks published on NAVER Webtoon, sampling 20k images only the works with permission from the original creators for research purposes.
Similar to the Painting dataset, we apply resolution filtering and resize the images to $512^2$.

To make the benchmark dataset, we select 20 objects from the personalization subject dataset proposed by \citet{ruiz2023dreambooth} for the Object domain.
Each object consists of 5-6 images.
When we conduct the main comparison (\tref{tab:comparison}), we use images resized to $512^2$, and for the arbitrary image analysis (\tref{tab:robustness}), we apply protection to the original resolution images.
For the Face domain, we followe \citet{van2023anti} and randomly sample 20 identities from VGGFace2~\cite{cao2018vggface2}, each identity consisting of 12 images resized to $512^2$.
The Painting dataset is compiled by randomly sampling (but not overlap to the training dataset) 20 artists from the WikiArt dataset, each artist represented by 10 artworks, all resized to $512^2$.
For the Cartoon domain, we randomly sample 20 Webtoon works (but also no overlap to the training dataset) from the Webtoon dataset.
In this dataset, we only use facial images of major cartoon characters by cropping and aligning all the images.
This reflects the common practice in cartoons and illustrations where characters are the main focus to both readers and artists.
In the main comparison we use $512^2$ images, while the arbitrary analysis maintains the original resolution.

The cartoon images used in the benchmark dataset are also permitted by the artists for research only purpose.
The list of cartoon works featured in this paper is as follows: \textit{\textless Yumi's Cells\textgreater}, \textit{\textless Maru is a Puppy\textgreater}, \textit{\textless Free Draw\textgreater}, \textit{\textless The Shape of Nightmare\textgreater}, \textit{\textless See You in My 19th Life\textgreater}, \textit{\textless Lookism\textgreater}, and \textit{\textless Love Revolution\textgreater}.

\tableSupplSubject{}
\tableSupplFace{}

\smallskip
\noindent\textbf{Baselines.}
\ours\ is compared with existing diffusion-based mimicry protection frameworks: AdvDM~\cite{liang2023adversarial}, PhotoGuard~\cite{salman2023raising}, Anti-DreamBooth (Anti-DB)~\cite{van2023anti}, Mist~\cite{liang2023mist}, Impasto~\cite{ahn2024imperceptible}, and SDST~\cite{xue2023toward}.
We follow the official settings for all the baselines by using their official codes.
However, for PhotoGuard, we use the target image of Mist~\cite{liang2023mist} by following the protocols of \citet{xue2023toward} while we use the Impasto's target image for the Impasto.

\smallskip
\noindent
\textbf{Evaluation.}
For the primary evaluation assessment, we use DISTS~\cite{ding2020image} to measure the invisibility of the protected image and FID~\cite{heusel2017gans} to evaluate the protection efficacy of the personalized diffusion methods.
In addition, in here, we also compare the protection frameworks using other metrics:
To evaluate invisibility, we include LPIPS\textsubscript{VGG}~\cite{zhang2018unreasonable} and AHIQ~\cite{lao2022attentions}.
Note that LPIPS\textsubscript{VGG} measures the perceptual distance in the VGG feature space, which is different from the feature space used in our perceptual map creation module (\ie AlexNet).
For evaluating protection efficacy, we use TOPIQ-NR~\cite{chen2024topiq} and QAlign~\cite{wu2023q}.

When preparing mimicry outputs, we personalize the diffusion models using the protected image with LoRA~\cite{hu2021lora} by default, employing default settings for all the personalization methods.
To generate outputs, we use different inference prompts than those used during training to simulate black-box caption scenarios~\cite{van2023anti}.
For example, we train the diffusion models with LoRA using captions of ``A painting in \textless\textit{sks}\textgreater\, style" prompt and perform inference with ``A painting of a house in \textless\textit{sks}\textgreater\, style".

\section{Discussions}

\textbf{Arbitrary Resolution.}
As shown in \tref{tab:robustness}, \ours\ can effectively handle arbitrary resolution images simply by resizing the MoP to match the resolution of an input image.
In contrast, the baseline method, PhotoGuard~\cite{salman2023raising}, suffers a much greater performance drop compared to ours.
There are two main factors that can impact protection performance for arbitrary resolution scenarios.
The first factor is the need to adjust the size of the perturbation to match the size of the image when injecting the perturbation.
In theory, iterative optimization methods do not face this issue.
However, in practice, due to the VRAM constraint of GPUs, images beyond (typically $1024^2$-px in a A100) a certain size must be split and protected in segments.
On the other hand, our model does not have memory issues, but since the pre-trained MoP is fixed as $512^2$-px, it requires resizing to fit an input image.
The second factor is the necessity to downsize the protected image due to resolution requirements that each diffusion models has (e.g., $512^2$ for SD v1, $768^2$ for SD v2, or $1024^2$ for SD-XL).

Our method appears to be affected by the information loss due to the two resize operations (one when applying MoP and the other during personalization).
Of course, we observed that all protection frameworks suffer some degree of protection efficacy reduction during personalization, and our model is no exception (e.g., FID: 220.3 to 204.0).
Such a reduction is due to the fine perturbations being lost during the downsize process (to match the resolution that of the diffusion models require).
In contrast, although PhotoGuard performs optimization in a full-resolution, its robustness in this aspect seems limited.
We suspect that the severe degradation occurs since the need to split the high-resolution images and more importantly the tendency of PhotoGuard to create very fine perturbations specific to each image make it particularly vulnerable to downsizing.
On the other hand, since our model learns perturbations that work well on average across the training data, the perturbations tend to be more coarse.
However, creating a more practically robust model requires further investigation and future work to fully reveal these aspects.

\smallskip
\noindent
\textbf{Assignment Function.}
In \ours, the assignment function $\mathcal{A}$ groups images as shown in \fref{fig:analysis_group}.
We observed that each perturbation has characteristic groupings of images.
For example, the first perturbation (top left in \fref{fig:analysis_group}) predominantly groups images of faces, portraits, or close-up shots of objects, with a notably dark background.
The second perturbation (top right) includes images with moderate scene complexity or texture.
The third perturbation (bottom left) selects the simplest images in the dataset, particularly cartoons or images with simple objects and minimal backgrounds in the Subject domain.
Finally, the fourth perturbation (bottom right) groups the most complex textured images.
For instance, in the subject domain, images with detailed textures like grassy backgrounds are mostly selected here, and in the face, cartoon, and painting domains, images with complex and detailed backgrounds or scenes are also grouped in this category.

\smallskip
\noindent
\textbf{Adaptive Targeted Protection.}
\tref{tab:appendix_target_image} shows a quantitative report of \fref{fig:analysis_target}.
Similarly, the performance in each test domain varies according to the pattern repetition of the target image.
For example, high repetition target images perform well in the Object domain but poorly in the Face and Cartoon domains.
This is because the Object domain mostly contains images with dense textures and complex scenes, which match well with the frequency of high repetition target images.
Conversely, the Cartoon domain, which typically has very flat textures, does not benefit from high repetition target images.
For the Face domain, even though it is similar to the Object domain in being natural photos, the backgrounds are blurred or monochromatic, and the textures of faces are frequently flat, differing significantly from objects.
For the low repetition target images, the opposite trend is observed.
Performance decreases in the Object domain but improves in the Face and Cartoon.
In the Painting domain, performance remains consistent across all target images.
This is because the artworks in the Painting domain exhibit a wide variety of textures and complexities, leading to an average performance across different target images.
For instance, the Painting domain dataset includes diverse images ranging from Van Gogh-style oil paintings to black and white sketch drawings.

\smallskip
\noindent
\textbf{Comparison at the Same Protection Strength.}
In the main paper, we compare \ours\ with other protection frameworks by adjusting the protection strength ($\eta$) to match the protection level across methods for a fairer comparison (\tref{tab:comparison}). The reason for this evaluation setup is that we observed different protection losses produce varying protection-invisibility trade-offs. Therefore, protection efficacy alone cannot be used to judge the superiority of a model. In \fref{fig:appendix_tradeoff}, we report the results by varying the protection strength and analyzing protection efficacy vs. invisibility. However, such benchmark requires significant computational resources, thus we adjust the protection strength to align the trade-off line for comparison.

Still, one might wonder about the performance when the protection strength is fixed for all methods. To answer this possible inquiry, we conduct a comparison with all protection frameworks by fixing the protection strength $\eta$ at 8, as shown in \tref{tab:comparison_same_budget}. In most cases, \ours\ demonstrates outstanding results in terms of invisibility while achieving moderate protection efficacy. This balance suggests that \ours\ performs exceptionally well when considering the protection-invisibility trade-off. In the Cartoon domain, Impasto~\cite{ahn2024imperceptible} shows better results in invisibility, but our method significantly outperforms Impasto in efficacy, which indicates a favorable trade-off trend.

\tableSupplPainting{}
\tableSupplWebtoon{}

\smallskip
\noindent
\textbf{Limitation \& Future Directions.}
As discussed in \sref{sec:conclusion}, all protection solutions, including our model, still need improvements in terms of invisibility.
However, as analyzed by \citet{ahn2024imperceptible}, to prevent mimicry, perturbations must be applied broadly across the image.
This characteristic makes it very challenging to achieve a high level of invisibility beyond a certain point.

Moreover, a common phenomenon across all models is that, even when the same budget of perturbation is applied to all images, the degree of protection varies from image to image.
Interestingly, if a specific model fails to protect a certain image, other models also tend to struggle with it.
This issue likely stems from the fact that all frameworks rely on texture or semantic losses (or both), thus they all share the same vulnerability.
A potential solution to this problem is to measure how difficult an image is to protect in advance and dynamically adjust the budget accordingly.
However, practically, it is challenging to determine the difficulty of protection before the personalization.
Therefore, predicting the protectability of input images and adjusting the intensity of perturbations adaptively is a worth a try research direction.
The another advantage of this approach is that easy-to-protect images can be given lower perturbation strength, naturally enhancing invisibility.

\figAppendixLatency{}
\figAppendixTradeoff{}

\section{Additional Results}
\label{sec:appendix_results}
\fref{fig:qual2} and \ref{fig:qual3} show additional qualitative comparisons of existing protection frameworks. In \tref{tab:appendix_comp_subject}, \ref{tab:appendix_comp_face}, \ref{tab:appendix_comp_painting}, and \ref{tab:appendix_comp_webtoon}, we evaluate the protection frameworks using the additional metrics.
In these comparisons, our model demonstrates better invisibility while showing average protection efficacy; similar protection performance with the texture loss model group.
Notably, models that primarily utilize texture loss (e.g., PhotoGuard, Mist, SDST, Impasto, and ours; although Mist and SDST utilize both losses, we observed that texture loss has a much higher influence) tend to exhibit relatively high invisibility. Conversely, AdvDM and Anti-DreamBooth, which use semantic loss, show higher protection efficacy in most of the protection efficacy measures.
This phenomenon highlights that the different losses exhibit varying trends. Understanding why these losses perform differently across metrics will be crucial for future advancements in protection loss design.

In \fref{fig:appendix_latency}, we compare the latency versus input image resolution on both CPU and GPU.
On the CPU (Apple silicon), \ours\ completes the protection in only 4 seconds, while all other methods require significantly more time; 15 minutes for PhotoGuard~\cite{salman2023raising} and 10 hours for Anti-DreamBooth~\cite{van2023anti}.
We also include the protection efficacy versus invisibility trade-off comparison for both the Object and Cartoon domains in \fref{fig:appendix_tradeoff}.
\ours\ demonstrates an improved trade-off curve in both domains.

\figQualTwo{}
\figQualThree{}